\documentclass[pdflatex,sn-vancouver]{sn-jnl}%
\usepackage[utf8]{inputenc}
\usepackage[english]{babel}
\usepackage{amsmath}
\usepackage{fancyhdr}
\usepackage{graphicx}
\usepackage{enumitem}
\usepackage{amsfonts}
\usepackage{pdflscape}
\usepackage{amsfonts}
\usepackage{caption}
\usepackage{amssymb}
\usepackage{subcaption}
\usepackage{xcolor}
\usepackage{color}
\usepackage{epstopdf}
\usepackage{float}
\usepackage{gensymb}
\usepackage{booktabs}
\usepackage{xurl}
\usepackage{flafter}
\usepackage{mathtools}
\usepackage{mathptmx}
\usepackage[numbers]{natbib}

\jyear{2022}%

\theoremstyle{thmstyleone}%

\theoremstyle{thmstyletwo}%

\theoremstyle{thmstylethree}%

\begin{document}

\title[Designing a Robust Low-Level Agnostic Controller for a Quadrotor with Actor-Critic Reinforcement Learning]{Designing a Robust Low-Level Agnostic Controller for a Quadrotor with Actor-Critic Reinforcement Learning}

\author*[1]{\fnm{Guilherme} \sur{Siqueira Eduardo}}\email{guiseduardo@gmail.com}

\author[1]{\fnm{Wouter} \sur{Caarls}}\email{wouter@ele.puc-rio.br}
\equalcont{These authors contributed equally to this work.}

\affil*[1]{\orgdiv{Laboratory of Intelligent Control (LCI)}, \orgname{Pontifical Catholic University of Rio de Janeiro (PUC-Rio)}, \orgaddress{\street{Rua Marqu\^{e}s de S\~{a}o Vicente, 225}, \city{Rio de Janeiro}, \postcode{22451-900}, \state{RJ}, \country{Brazil}}}

\abstract{\textbf{Purpose:} Real-life applications using quadrotors introduce a number of disturbances and time-varying properties that pose a challenge to flight controllers. We observed that, when a quadrotor is tasked with picking up and dropping a payload, traditional PID and RL-based controllers found in literature struggle to maintain flight after the vehicle changes its dynamics due to interaction with this external object.

\textbf{Methods:} In this work, we introduce domain randomization during the training phase of a low-level waypoint guidance controller based on Soft Actor-Critic. The resulting controller is evaluated on the proposed payload pick up and drop task with added disturbances that emulate real-life operation of the vehicle.

\textbf{Results \& Conclusion:} We show that, by introducing a certain degree of uncertainty in quadrotor dynamics during training, we can obtain a controller that is capable to perform the proposed task using a larger variation of quadrotor parameters. Additionally, the RL-based controller outperforms a traditional positional PID controller with optimized gains in this task, while remaining agnostic to different simulation parameters.}


\maketitle

\section{Introduction}\label{sec:intro}

Recent advances in computational power and in the efficiency of machine learning algorithms have made it possible to use neural networks for various types of complex applications that require autonomy and comprehensive perception of the environment. These advances have made it faster to calculate the output of these networks even in embedded controllers, enabling novel approaches to control stability, navigation, perception, \textit{et cetera}, in mobile robots \cite{lcdet, lillicrap2015continuous}.

The quadrotor is an example of such a robot that is able to move in 3-dimensional space using four rotors, each consisting of a motor and a propeller. This vehicle is open loop unstable during flight, requiring constant actuation to hover and move around without flipping or falling to the ground \cite{robustgarcia}. Conventionally, underlying PID controllers are used to provide closed loop stability to the quadrotor so that higher level controllers can focus on more difficult tasks \cite{lambert2019low}. The issue with this kind of controller is that it assumes that the system it tries to control is linear and time-invariant, while the quadrotor has non-linear dynamics and has some time-variant variables, such as battery voltage, motor performance (due to wear or temperature), or even mass, if the vehicle is tasked to interact with a payload.

%%%

To deal with uncertainties inherent to the quadrotor model, classical and modern control solutions have been employed, such as a robust PID design strategy and adaptive controllers. A model reference adaptive control strategy has been used to augment a baseline trajectory controller in real flight, accounting successfully for severe loss-of-thrust failure \cite{adaptivedydek}. A similar strategy has been employed for the fully actuated quadrotor subsystem, with a sliding model controller for the internal dynamics instead, where the plant parameters are fully unknown during controller design \cite{emran2014robust}.

In case of robust nonlinear PID design, a strategy consisting in the H-infinity norm has been investigated for the complete nonlinear quadrotor model \cite{robustgarcia}. This strategy is employed to tackle similar issues with uncertainties not present in the quadrotor model, as well as delays or variations in measurements. However, this controller has struggled with larger variations of the reference signals. A cascaded strategy has been investigated using feedback linearization coupled with a PD controller for translational control with a backstepping controller for controlling the internal dynamics \cite{mian2008modeling}, employed in quasi-stationary flights with no disturbances. For more complex disturbance input settings, a cascaded strategy has been investigated using a backstepping controller with integral action for trajectory tracking control with an H-infinity controller for the internal dynamics \cite{raffo2015robust}. This combination has been shown to cope with saturated control inputs and crossed inertia terms in the moment of inertia matrix. While robust and adaptive nonlinear control has been demonstrated to be able to handle a number of uncertainties, the design process of such controllers requires complex adaption laws, information about all system states, and extensive knowledge of vehicle parameters \cite{mo2019nonlinear} (except \cite{emran2014robust}, which considers the vehicle parameters to be unknown).

Machine learning (ML) methods have been seeing increased use for real-time robot control as a result of increased computational power available in embedded computers, especially the employment of artificial neural networks (ANNs) as a strategy for devising a control scheme for quadrotors. In an extensive comparative study regarding quadrotor controllers, Mo and Farid \cite{mo2019nonlinear} acknowledge that most of the work in this field is restricted to "high-level" control approaches, where ML algorithms rely on an existing underlying low-level controller. However, there are recent works that attempt establishing a direct sensor-actuator control scheme using ANNs \cite{beckerehmck2020learning, Hwangbo2017, koning2020low}.

These methods owe their popularity not only to the increased available embedded computational power, but to significant improvements in control performance that can be achieved by the unique structure of ANN-based controllers and of on-line learning algorithms. These control architectures have been able to cope with progressively more complex uncertainties in the quadrotor dynamics, in addition to the ones considered in the aforementioned robust nonlinear controllers (e.g. sensor noise, sudden changes in mass, loss of thrust), as well as improved adaption for uncertain model parameters, completely disregarding the need for an exact vehicle model \cite{mo2019nonlinear}. Besides the robustness aspect of ANN-based controllers, reinforcement learning and other machine learning methods do not require extensive prior proficiency in modelling and controlling the system in question, when in fact the controller is able to learn and adjust for the system dynamics automatically, which speeds up the controller design and deployment process, especially for larger and more intricate systems.

%%%

As universal function approximators \cite{annbook3}, ANNs can replace traditional methods of controller design by fitting a function that describes the relationship between the feedback error and control actions for the controlled system. ANNs employ non-linear features into their hidden layers \cite{Goodfellow-et-al-2016}. As such, they are able to model the non-linear aspects of the quadrotor, conceivably, even, if these aspects are not directly observable, for example, using coarse localization of scene features in a GPS-denied environment \cite{myself_master}. Neural networks with multiple neuron layers built on top of each other are regarded as "deep" neural networks, thus establishing the process of deep learning \cite{Goodfellow-et-al-2016}.

Generally, a neural network is obtained via supervised learning, where training data consists of a paired set of inputs and outputs for the desired function to approximate \cite{Goodfellow-et-al-2016}. However, there is no way to obtain a training set for a low-level quadrotor controller without a fully developed knowledgeable controller. In this case, we investigate a Reinforcement Learning (RL)-based approach: instead of finding the desired control output for the network, we need to design a reward function that describes the desired behaviour of the quadrotor, which should be trivial \cite{sutton}, as we could, for example, attribute rewards for successfully hovering and costs to destabilization.

%%%%%%%%%%

Previous work on employing RL on quadrotor control, such as \cite{koning2020low, koch2018reinforcement, lambert2019low, Hwangbo2017, barros2020unicamp}, largely focus on controlling a single type of quadrotor, often because of restrictions imposed by the simulations used or the employment of real vehicles. This work, instead, focuses on a generalized parameter-agnostic approach for designing quadrotor navigation controllers using reinforcement learning techniques, outlining the capabilities and limitations of such controllers in a variety of tasks, stressing the requirement for such controllers to be stable, accurate and robust to different quadrotor parameters and to sudden changes of these parameters, without the need to re-train the controller model. We show that it is possible to design a low-level RL controller for a quadrotor that is, at the same time, stable and accurate, and robust to a range of parameters, time varying circumstances and disturbances.

To achieve this objective, this work sets to investigate the integration of domain randomization into a Deep Reinforcement Learning (DRL)-based quadrotor controller. In this approach, the controller is trained in simulations with varying features in the vehicle's basic structure, here referred to as vehicle parameters, such as mass and propeller size. This approach has been shown to increase the accuracy of sim-to-real transfer of learned policies \cite{peng2018sim} and, in this work, additionally, it is employed to prevent overfitting of an RL-based controller to a specific quadrotor model, by forcing it to generalize behaviour for unpredictable parameters. Compared to a controller trained using the proposed method, we show that a controller designed to maximize performance on a single set of vehicle parameters underperforms when exposed to widely different simulator parameters.

To accomplish that, we employ Soft Actor-Critic (SAC) \cite{haarnoja2018soft}, a recent DRL algorithm, that includes entropy maximization during training, which, among other advantages, encourages many alternatives for control, where possible, enabling the adaptability necessary to maintain optimal behaviour in such a range of simulation parameters. This RL algorithm is used to train a controller that directly maps sensor data into motor inputs. For the sake of comparison and evaluation, we develop two other controllers: one that controls only the desired velocity, working together with a low-level PID controller that provides the corresponding motor inputs, emulating a common control cascade found in literature \cite{lambert2019low, mo2019nonlinear}, and a full traditional PID controller, which composes the core of \textit{Ardupilot}\footnote{$<$\url{https://ardupilot.org/}$>$ Accessed: 24 Sep. 2022}, a widely used open-source flight controller for a multitude of aerial vehicles. For fairness of comparison, both PID controllers proposed for benchmarking undergo gains tuning on the same range of quadrotor parameters as used for training other controllers, via a CMA evolutionary strategy search algorithm \cite{cmaes}, in order to ensure that these controllers have close to optimal performance under the proposed conditions.

Further, we propose a payload pick up and drop course that takes full advantage of the agnosticism of the proposed controller. When picking up or dropping a payload, the quadrotor mass changes suddenly and significantly, requiring the controller to be able to not only maintain flight stability, but must present equal performance in carrying out the remainder of the task. During this course, the vehicle is also subjected to additional disturbances, emulating two common phenomena found in real-world scenarios: sensor noise and motor command delay. A robust controller that is able to account for a wide range of quadrotor parameters is not only useful to account for the mentioned disturbances, but also to smooth out the transition from simulation to reality, one of the challenges of low-level DRL controllers \cite{koch2018reinforcement}, as we can train the model to expect the uncertainties of a real quadrotor.

In summary, we show that:
\begin{itemize}
\item An RL-based based controller can outperform a PID controller in a scenario where vehicle dynamics change mid-flight (no opportunity to re-tune PID gains);
\item A randomized training environment is key for an increased robustness of the RL-based controller. We compare two controllers with the same architecture and the same RL algorithm (SAC), one trained with environment randomization and another one without; and
\item An RL-based controller doesn't need an underlying fast PID controller to perform well, a low-level controller performs just as well or better in specific tasks. We compare two controllers trained with the same method, but with two different end goals: actuating directly into motor commands and providing setpoint velocities to an underlying PID controller.
\end{itemize}

\section{Background}\label{sec:background} % (Quadrotors, RL, DRL, RL for quadrotors)

Reinforcement learning is an area of machine learning used to obtain a control policy that maximizes a reward signal $R$ over time, without any other form of prior knowledge made available. This is accomplished by predicting the expected return starting from a state $s$ taking action $a$ by approximating the state-action value function $Q(s,a)$. Then, this return is maximized by following a suitable policy $a=\mu(s)$. In most modern applications, the state-action value is represented by a parameterized function differentiable with respect to its parameters $\mathbf{w}$, thus being denoted as $Q_{\mathbf{w}}(s,a)$ \cite{sutton}.

Actor-critic methods are a popular form of implementation of RL algorithms for continuous actions spaces. In these methods, the control policy is represented by a parameterized function that can be differentiated with respect to these parameters ($\mathbf{v}$), thus being denoted as $\mu_{\mathbf{v}}(s)$ for a deterministic policy, and $\pi_{\mathbf{v}}(s)$ for a stochastic policy. This control policy is updated by calculating its gradient, with respect to its parameters, in the direction with the highest cumulative rewards. Thus, these methods are called policy gradient methods \cite{silver14}.

Deep Q-Network (DQN)-based Actor-Critic methods are a popular form of policy gradient algorithms \cite{silver14}, in which a differentiable control policy (the actor) is improved based on the judgment of a value function (the critic). The simplest policy gradient computed by this method is:

\begin{equation}
\nabla_\theta Q_{\mathbf{w}}(s,\mu_{\mathbf{v}}(s)),
\end{equation}\label{eq:2pol_gradient}

\noindent which updates the policy parameters via a deterministic policy gradient (DPG) ascent \cite{silver14}, as to maximize the expected return of the control policy over the states.

Deep Deterministic Policy Gradients (DDPG) \cite{lillicrap2015continuous}, Twin Delayed DDPG (TD3) \cite{fujimoto2018addressing} and Soft Actor-Critic (SAC) \cite{haarnoja2018soft}, the latter being employed throughout this work, are widely used actor-critic methods that introduce a number of features to the base Deterministic Policy Gradient DPG \cite{silver14} algorithm, addressing a number of common problems. In short, these improvements add up to:

\begin{itemize}
\item Improving learning stability, which is usually brittle, by sampling $Q$-value estimates from target $Q$-networks \cite{lillicrap2015continuous} and sampling control actions from a target policy network \cite{fujimoto2018addressing}, both of which track their corresponding parent networks via Polyak averaging;
\item Addressing distribution shift over time and sample efficiency in on-line learning environments, taking advantage of off-policy properties of the algorithms to employ a replay buffer \cite{silver14, lillicrap2015continuous};
\item Adding regularization in policy updates (soft policy updates) \cite{fujimoto2018addressing}; and
\item Addressing overestimation bias of the $Q$-value function, by sampling the minimum value of two independent $Q$-networks (double $Q$-learning) \cite{fujimoto2018addressing}.
\end{itemize}

The SAC algorithm was chosen for a number of reasons, the most important being that, as an off-policy algorithm, with automatically adjusted exploration, it benefits from greater sample efficiency than previous actor-critic algorithms, leading to faster learning. Furthermore, it requires less hyperparameter tuning by the user, and we expect the maximum entropy RL feature, presented by the algorithm, to help the control policy to easily adapt to different quadrotor parameters.

%%%%%%%%%%%%%%%%%%%%%%%%%%%%%%%%%%%%%%%%%%%%%%%%%%%%%%%%%%%

\subsection{Soft Actor-Critic}

One of the most important features of the SAC algorithm is that it automatically adjusts exploration of the agent by enforcing a maximum entropy target, in other words, a target of maximum "randomness" of the control policy, resulting in a control policy that is able to devise multiple routes to an objective. A version of this algorithm by the same author \cite{haarnoja2019soft} goes further and automatically adjusts the reward scale of the entropy target, denominated the "temperature". By using this method, not only does it become unnecessary to tune exploration, but we can also expect the agent to automatically seek improvement in subspaces where there is a low confidence in $Q$-value prediction, ensuring resulting control policies closer to optimal. The enforced entropy objective further speeds up learning compared to previous actor-critic iterations, a crucial feature if one wishes to employ RL on a real-life quadrotor \cite{haarnoja2018soft}.

\subsubsection{Maximum entropy objective}

To enforce entropy maximization, the reward term is augmented with an entropy term $\mathcal{H}(\pi_{\mathbf{v}}(\cdot\vert s))$, which represents the entropy of the distribution over actions $\pi_{\mathbf{v}}(\cdot\vert s)$ of the stochastic control policy. The augmented reward becomes, at time $t$:

\begin{equation}\label{eq:2sac_rew}
R_t = r(s_t,a_t,s_{t+1}) + \tau\mathcal{H}(\pi_{\mathbf{v}}(\cdot\vert s_t)),
\end{equation}

\noindent where the temperature, $\tau$\footnote{Temperature is denoted as $\alpha$ by \cite{haarnoja2018soft}. Here we use $\tau$ to not confuse with propeller pitch ($\alpha$).}, dictates the scale of the effect of the entropy term in the value function and in the stochastic policy. The entropy is a measurement of the "randomness" of a variable. Using samples from the policy, we can estimate the entropy by:

\begin{equation}\label{eq:2sac_Hbecomeslog}
\mathcal{H}(\pi_{\mathbf{v}}(\cdot\vert s)) = -\log\pi_{\mathbf{v}}(a\vert s), \qquad a\sim\pi_{\mathbf{v}}(\cdot\vert s) .
\end{equation}

\subsubsection{Automated entropy adjustment}

The temperature $\tau$ defines the scale of the influence of the entropy term in the reward function. A high temperature causes the algorithm to approximate a high-entropy low-return control policy, while a low temperature causes the opposite. This complex trade-off has different effects with different training environments, usually requiring manual tuning of the temperature for each environment. Further, the temperature requirement depends on the policy, which changes during the learning process. A solution consists in formulating a different maximum entropy reinforcement learning objective, where the entropy is treated as a constraint \cite{haarnoja2019soft}.

In this formulation, the average entropy of the policy is set as a constraint, $h$, and the temperature $\tau$ is adjusted by minimizing the loss function:

\begin{equation}\label{eq:2temp_update}
\mathcal{L}(\tau) = -\ln{\tau} (\mathcal{H}(\pi(\cdot\vert s_t)) + h)
\end{equation}

\noindent after performing the value update and policy improvement steps.

As a result, the complex interplay between maximum entropy objective and maximum expected return from the environment is controlled in accordance to a minimum expected entropy constraint, adjusting the scale of "stochasticity" according to the needs of the environment and the exploration requirements \cite{haarnoja2019soft}. In other words, if the policy entropy is below the constraint, $\tau$ increases, emphasizing the maximum entropy objective. On the other hand, if the policy entropy is higher than the minimum entropy constraint, $\tau$ decreases, emphasizing the maximum expected return.

%%%%%%%%%%%%%%%%%%%%%%%%%%%%%%%%%%%%%%%%%%%%%%%%%%%%%%%%%%%

\subsection{Quadrotor dynamics}

The quadrotor, also known as quadcopter, is a kind of unmanned aerial vehicle (UAV, also commonly known as drone) that generates lift using 4 rotors, one at the extremity of each of its arms. Most quadrotors have these rotors distributed symmetrically in a cross ($\times$) shape or in a plus ($+$) shape, which is employed throughout this work.

\begin{figure}[tbp]
\centering
\includegraphics[width=.8\linewidth]{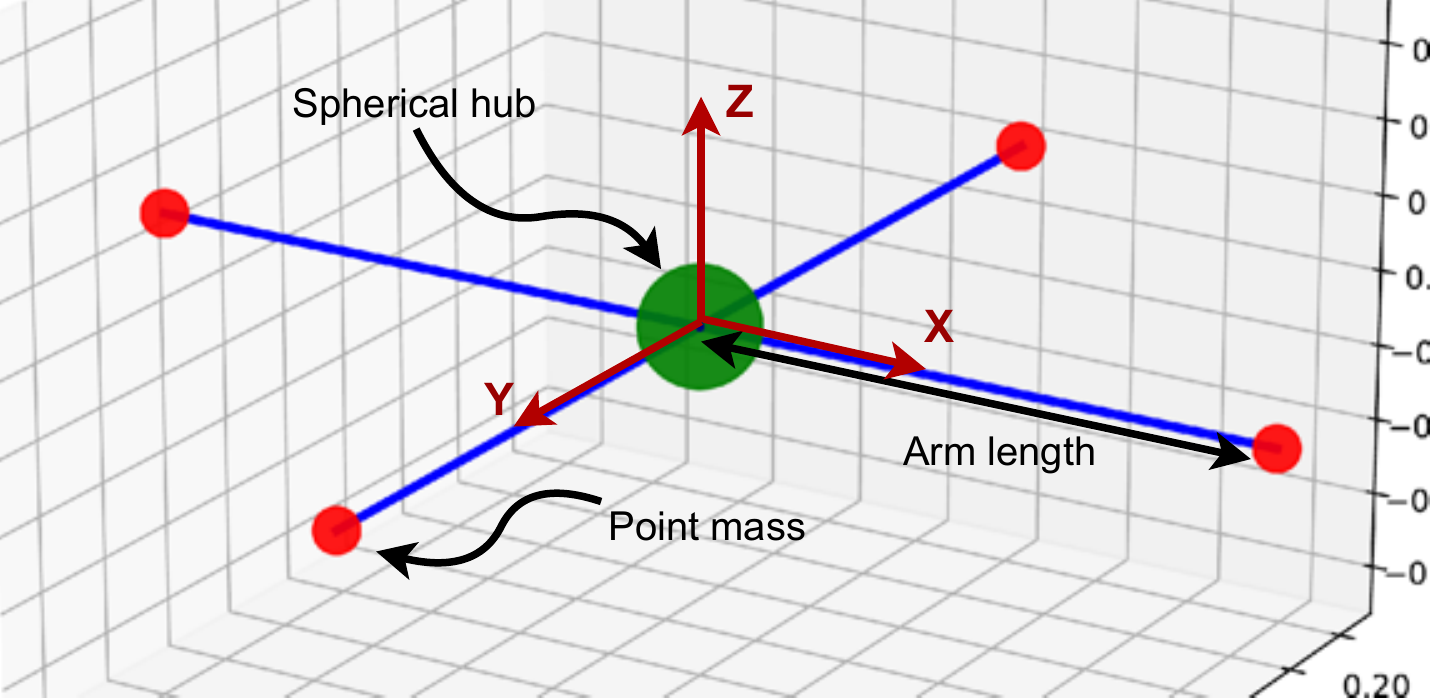}
\caption{Simplified model of the quadrotor used in simulation.}
\label{fig:params_quad3d}
\end{figure}

Model training and experiments are performed with a simulated quadrotor based on the dynamic model of a simplified structure illustrated by Fig.~\ref{fig:params_quad3d}. The central body of the quadrotor, or the hub, which houses the battery and all the electronic components of the vehicle, as well as any extra payload, is approximated by a sphere. Each of its arms is approximated by a thin rod with no mass and its rotors approximated by point masses \cite{Beard2008QuadrotorDA}.

While the local and inertia reference frames most used for quadcopter dynamics modelling is NED (north-east-down), where the axes X, Y and Z point to, respectively, front, right and down, the formulation used in this work uses the NEU reference frame (north-east-up), so the local reference frame is X pointing to the front, Y pointing to the right and Z pointing up, as indicated by the diagram.

%%%
The thrust $T_i$ produced by each rotor results from the relationship between rotor velocity and propeller parameters described by:

% eq. propeller
% thrust = 4.392e-8 * speed * (diam ^3.5)/(sqrt(pitch))*(4.23e-4 * speed * pitch)
\begin{equation}\label{eq:2thrust}
T_i = 4.392\times10^{-8} \; \Omega \frac{d^{3.5}}{\sqrt{\alpha}}\; (4.23\times10^{-4}\; \Omega\; \alpha),
\end{equation}

\noindent based on the aerodynamics of a dual-blade propeller with diameter $d$ and angle of attack defined by a "pitch" ($\alpha$) calculation\footnote{Staples G (2014) Propeller Static \& Dynamic Thrust Calculation. $<$\url{https://www.electricrcaircraftguy.com/2014/04/propeller-static-dynamic-thrust-equation-background.html}$>$ Accessed: 12 Mar. 2021}. In this model, the propeller pitch is the theoretical travel distance of the propeller after one full revolution, assuming that the leading edge of the blades have an angle of attack of $0^\circ$ with respect to the airflow (no slip). Both propeller parameters are measured, in this case, in inches. In the same equation, $\Omega$ is the desired rotor velocity (in RPM, revolutions-per-minute), the direct input of this model, assuming the electric motor subsystem dynamics are negligible with respect to the overall system.

The set of differential equations that describe the dynamics of the quadrotor in the local and inertial frames are described by\footnote{Gibiansky A (2012) Quadcopter Dynamics, Simulation, and Control. $<$\url{https://andrew.gibiansky.com/downloads/pdf/Quadcopter\%20Dynamics,\%20Simulation,\%20and\%20Control.pdf}$>$ Accessed: 12 Mar. 2021}:

\begin{equation}\label{eq:2model_complete}
\begin{split}
\dot{\nu} & = -\omega\times\nu + R^\intercal g + F_p/m \\
\dot{\omega} & = -I^{-1}\omega\times I\omega + I^{-1}M_p \\
\dot{p} & = R\nu \\
\dot{\Phi} & = S\omega ,
\end{split}
\end{equation}

\noindent where $R$ is the rotation matrix from the local reference frame to the inertial frame, $S$ is the representation Jacobian which describes the relationship between the quadrotor angular velocities $\omega$, in the local frame, to the angular rates $\dot{\Phi}$, in the inertial frame, and $I$ is the moment of inertia of the simplified quadrotor model. The input of the model, $\Omega$, translates into thrust forces $T_i$ exerted in the edge of each arm via the thrust model in Eq.~(\ref{eq:2thrust}). The resulting force and momentum exerted in the airframe as a result of thrust are represented by, respectively, $F_p$ and $M_p$.

The observation array used as input to the reinforcement learning controllers is the same state space array $[\nu, \omega, p ,\Phi]$, where:

\begin{itemize}
\item $\nu$ and $\omega$ are, respectively, the linear and angular velocities of the quadrotor in the local frame, and
\item $p$ and $\Phi$ are, respectively, the quadrotor position in the setpoint frame, also referred as X, Y and Z positions, and its attitude in the setpoint frame, which is its angular position around axes X (roll), Y (pitch) and Z (yaw).
\end{itemize}

%%%%%%%%%%%%%%%%%%%%%%%%%%%%%%%%%%%%%%%%%%%%%%%%%%%%%%%%%%%

\subsubsection{Linearization}

PID controllers are originally designed for controlling linear time-invariant systems. Because most practical systems are non-linear, such as the quadrotor, some adaptations are needed for the controller and/or for the plant. A straightforward adaptation commonly used is a linear approximation by first order Taylor series, where the system is modelled as a linear function around a reference point. For quadrotors, this reference is the hover point, where all velocities and attitude angles are zero \cite{dronex4}.

In the linearized model of the quadrotor, movement in each of the 6 degrees of freedom become uncoupled double integrator subsystems, allowing to design a controller for each subsystem separately as a single input single output (SISO) plant \cite{dronex4}. Then, the desired control action, $\delta$, calculated as throttle, roll, pitch and yaw torques, can be allocated directly to individual motor velocities $\Omega$, via a control allocation matrix $\Delta$:

\begin{equation}\label{eq:2controlallocationp}
\Omega = \left[ \begin{array}{cccc}
1 & 1 & 0 & 1 \\
1 & 0 & 1 & -1 \\
1 & -1 & 0 & 1 \\
1 & 0 & -1 & -1 \end{array} \right] \delta ,
\end{equation}

\noindent specific for the plus-shaped quadrotor.

\subsection{Related work}

Deep reinforcement learning (DRL) control applications aim to develop a controller that fills in the same positions as classic PID and similar controllers, as well as more complex ones. Control tasks are subdivided in two levels: high-level control and low-level control, in which a controller can fulfill either or both.

Low-level controllers act directly in the quadrotor actuators, as throttle-roll-pitch-yaw (throttle-RPY) forces and torques, which are allocated to each motor, or controlling motors directly. These controllers are usually required to be fast and robust, as they are essential for keeping the vehicle stable.

High-level controllers, on the other hand, do not interact directly with actuators. They, instead, map high level sensory information into commands that are passed down onto low-level controllers. Intelligent and experimental control tasks often rely on a pre-existing stable underlying on-board controller that manages the low-level control loop \cite{lambert2019low, mo2019nonlinear, lee2021sacher}. This is done often because the underlying dynamics of quadrotors are notoriously difficult to control, especially if its parameters are unknown or difficult to accurately measure. In addition, most commercial quadrotors already contain an underlying controller of this kind.

Low-level control tasks are traditionally fulfilled by PID controllers. However, recent work has been conducted to employ RL-trained controllers into this role. The controllers developed in \cite{beckerehmck2020learning, Hwangbo2017, koning2020low, barros2020unicamp} not only successfully fulfill this role, but also can fulfill both low and high level control loops at the same time, showing that learning-based controllers are able to grasp the dynamics of the quadrotor by themselves.

Model-based RL (MBRL) methods have been employed in training with real vehicles in \cite{lambert2019low} and \cite{beckerehmck2020learning}. This RL strategy consists in producing, using real data, a reference model of the system which is controlled, in this case, the quadrotor. In this approach, the available training samples are complemented using samples from the reference model, improving sample efficiency and reducing the amount of observations needed from the real system, reducing wear, in case of a physical system, and speeding up the learning process. In \cite{beckerehmck2020learning}, the reference model is a latent state-space model constructed from observational data, while in \cite{lambert2019low}, the reference model is a model predictive controller (MPC) constructed using trajectory samples. Both works perform experiments with physical quadrotors. In case of \cite{lambert2019low}, a \textit{Crazyflie} quadrotor is used to learn a hover controller, while \cite{beckerehmck2020learning}, uses a custom-made vehicle to learn a waypoint tracking controller. According to the findings by these works, working on real hardware is a challenge for this kind of learning because sensors often present noise and drift in their measurements, interfering with predictions. Nonetheless, both works found that MBRL is able to cope with these disturbances. The controllers developed using these methods can achieve very satisfactory robustness and performance targets for the particular quadrotor employed in learning. However, these controllers do not explicitly account for uncertainties emerging from a sudden difference in the dynamics of the quadrotor with respect to the dynamics observed during training and during the construction of the reference model. This differences might occur by simply loading the controller on another quadrotor or might occur mid-flight, due to interaction with the environment or damage to components.

Model-free methods have also been investigated for the presented task. Previous works, such as \cite{Hwangbo2017}, have successfully been able to develop a low-level controller using a variation of classic RL algorithms within simulated environments, capable of waypoint tracking and fast vehicle stabilization. The RL algorithm developed in the mentioned work is an adaptation of deterministic policy optimization (or gradient, such as DDPG) with a natural gradient descent, including added optimizations, under the assumption that the training environment is deterministic, resulting in a very sample-efficient, yet model-free, RL algorithm.

A deterministic environment indicate that the same actions under a certain state $s$ will always result in the same state $s'$, otherwise the system is considered stochastic. While the system considered by \cite{Hwangbo2017} is deterministic, an actual quadrotor might present certain stochastic state transitions due to time-varying properties already discussed, which require automatic adaptation by the controller. Further, in our work, stochasticity in the quadrotor model is made even more present by the introduction of uncertainty of vehicle parameters, such as mass and size. In this case, a stochastic policy gradient algorithm might be more appropriate, as, according to the authors, these methods tend to broader classes of problems \cite{Hwangbo2017}.

More recently, actor-critic algorithms have been showing good learning performance, while being easier to implement and requiring fewer hyperparameter adjustments. Some examples of recent work using such algorithms are \cite{koning2020low},\cite{koch2018reinforcement} and \cite{barros2020unicamp}. In \cite{koning2020low}, DDPG and TD3 are investigated to train a deterministic quadrotor controller for hover control and waypoint tracking, as well as some variations on the reward function. While adequate results were found by the authors, we found that training tends to diverge using these algorithms, even when using a deterministic model of the quadrotor.

In the case of \cite{koch2018reinforcement}, DDPG, TRPO and PPO are investigated for learning an attitude controller using accurate simulated quadrotor models. This controller operates in cascade with a higher-level PID controller used to control for translational movement. The authors describes three open challenges in RL for the task at hand (but also applicable for many RL robot control tasks): precision and accuracy; robustness and adaptation; and reward engineering. In their work, the first challenge, precision and accuracy, is explicitly addressed, with experiments resulting in high-precision controllers, based on very strict control policies, which the authors claims to be the appropriate approach to such a time-sensitive task as the attitude control. While not so explicit, other previous works tend to lean towards the same objective, favoring accuracy for a deterministic quadrotor model over robustness to stochastic factors.

Another example of such case is \cite{barros2020unicamp}, where the authors employ SAC, the same algorithm that is proposed for our work, to train a simulated quadrotor based on a real model to perform waypoint tracking. The controller is trained on a static target and, then, tested on a moving target. The proposed evaluation method is unable to prove the robustness of the controller because no additional disturbances are added to the model, in other words, the evaluation conditions are the same as training conditions: it doesn't matter for the controller if the target waypoint has changed over time, because, in Markovian stochastic processes, the probability distribution, in a given environment, of state transitions and rewards given a state-action pair, $p(s',r \vert s,a)$, is independent of all previous states visited and actions taken \cite{sutton}. It is under this assumption that, further in Sect.~\ref{sec:llc-test}, we introduce waypoint guidance to manage long-distance waypoints. Furthermore, the authors encounter a significative steady-state error in static waypoint tracking, even in training conditions. We show in our work how this error might be the result of overfitting to a single quadrotor model, a challenge which is tackled by the proposed randomized environment training strategy.

Considering the open challenges in RL put forth by \cite{koch2018reinforcement}, our work, unlike other previous works, aims at the second challenge, robustness and adaptation, when designing the robust waypoint tracker. We show that it is possible to design a low-level RL controller for a quadrotor that is, at the same time, stable, accurate and robust to a range of parameters and time varying circumstances, which were not present in the investigated previous works. This is achieved by leveraging the entropy enforcement introduced by the novel Soft Actor-Critic algorithm. In fact, as previously mentioned, the introduction of uncertainty in quadrotor parameters leads the DDPG and TD3 to diverge in our training environment, while learning tends to be more robust using SAC.

%%%%%%%%%%%%%%%%%%%%%%%%%%%%%%%%%%%%%%%%%%%%%%%%%%%%%%%%%%%

\section{Methods}\label{sec:methodology} % (Particular algorithms and adaptations used)

%%%%%%%%%%%%%%%%%%%%%%%%%%%%%%%%%%%%%%%%%%%%%%%%%%%%%

The task of this work is to develop a full-stack quadrotor controller using RL. In order to carry out evaluation and experimentation, two additional controllers need to be developed: a standard PID controller and a combination of both PID and RL (referred to as cascade controller). At first, the controller should take in sensor information data and translate directly into motor commands. While this "direct connection" is possible using a fully learned controller, it is not trivial to do so using PID, for example. \textit{Ardupilot}'s strategy for position control consists of a cascade of 4 controllers: a P controller to convert position errors to velocity setpoints, combined with a PID controller to convert velocity errors into attitude setpoints\footnote{$<$\url{https://ardupilot.org/dev/docs/code-overview-copter-poscontrol-and-navigation.html}$>$ Accessed: 24 Sep. 2022}, which are, then, cascaded into a P controller that converts attitude errors into attitude rate setpoints, combined with a PID controller that converts attitude rate errors into motor commands\footnote{$<$\url{https://ardupilot.org/dev/docs/apmcopter-programming-attitude-control-2.html}$>$ Accessed: 24 Sep. 2022}. In this work, considering the tuning method chosen for the full controller, the mentioned strategy is simplified by grouping the P controllers together with their subsequent PID controllers, resulting in two cascaded control subsystems: a PID controller that converts position errors directly to attitude setpoints (simply referred as outer or external PID), which cascades into a PID controller that converts attitude errors to motor commands (simply referred as inner or internal PID) \cite{dronex4}.

\begin{figure}[htb]
\centering
\includegraphics[width=.8\linewidth]{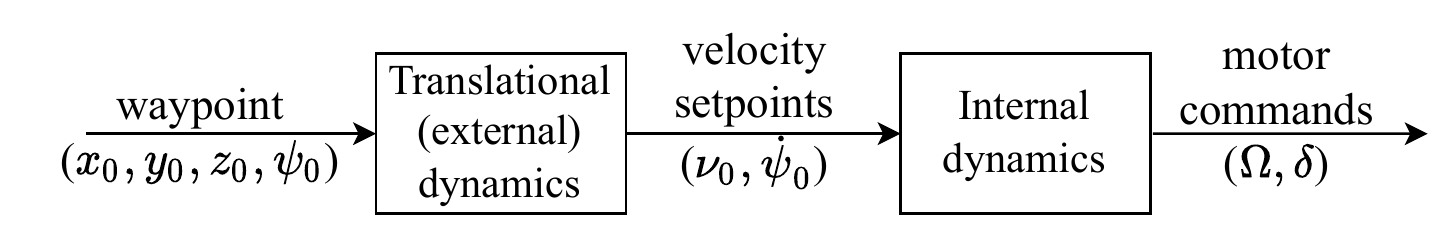}
\caption{Control stack for the cascade controller, used to benchmark the main RL-based controller.}
\label{fig:3diag_intro}
\end{figure}

For the cascade controller to be employed in benchmarking, based on what has been developed in literature, the PID controller needs to be modified in order to accommodate the high-level RL controller. This is done by further dividing the control stack into an additional subsystem: the translational dynamics of the quadrotor in the inertial frame, controlled using the high-level RL controller, which cascades into an internal subsystem that represents the dynamics of the quadrotor in the local frame. This internal subsystem is controlled by a modified version of the PID controller developed, which takes velocity setpoints directly as inputs, instead of the quadrotor pose. The relationship between subsystems is described in Fig.~\ref{fig:3diag_intro}.

To ensure the parameter agnosticism of the developed controllers, the training environment of the RL-based controllers is randomized with respect to the quadrotor mass, hub size, arm length and propeller diameter, with no way for the controller to know these beforehand. For fairness of comparison, the PID-based controllers receive the same treatment: optimized gains are found via a CMA evolutionary strategy search algorithm \cite{cmaes}, also in randomized environments, to ensure that these controllers have close to optimal performance under the proposed conditions. Besides the randomization of vehicle parameters, the environments have motor saturation as the only disturbances, while additional disturbances, namely sensor noise and motor command delay, are added only in experiment environments. To benchmark the efficacy of environment randomization in the robustness of an RL controller, a second fully learned controller is produced with this feature withheld from the training environment.

%%%%%%%%%%%%%%%%%%%%%%%%%%%%%%%%%%%%%%%%%%%%%%%%%%%%%
%%%%%%%%%%%%%%%%%%%%%%%%%%%%%%%%%%%%%%%%%%%%%%%%%%%%%
%%%%%%%%%%%%%%%%%%%%%%%%%%%%%%%%%%%%%%%%%%%%%%%%%%%%%
%%%%%%%%%%%%%%%%%%%%%%%%%%%%%%%%%%%%%%%%%%%%%%%%%%%%%

\subsection{Controller integration}\label{sec:controller_integration}

In this subsection, detailed description of the control loops of each of the developed controllers is provided, with respect to the proposed subsystem division of Fig.~\ref{fig:3diag_intro}.

\begin{figure}[tbh]
\centering
\includegraphics[width=.8\linewidth]{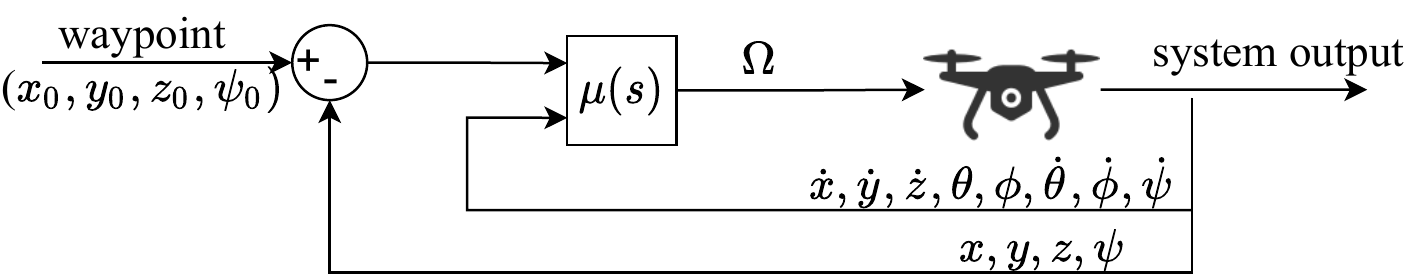}
\caption{Control diagram for the fully learned controller. The $\mu(s)$ block represents the ANN trained using the SAC algorithm.}
\label{fig:3diag_fullrl}
\end{figure}

The learning-based controllers are guided to a waypoint by minimizing the position error of the quadrotor with respect to this waypoint. The diagram in Fig.~\ref{fig:3diag_fullrl} illustrates how the learned control policy ($\mu(s)$) encompasses the full controller, which works in a simple feedback loop, where this $\mu(s)$ block is an implicit ANN structure.

In the diagram, the error tuple that we are aiming to minimize is the 3-dimensional position ($x,y,z$) and the vehicle yaw ($\psi$). At the same time, this controller receives the additional IMU data required to minimize attitude angles and better observe the current state of the agent. The IMU provides, additionally, the quadrotor roll ($\theta$), pitch ($\phi$), linear velocities ($\dot{x}, \dot{y}, \dot{z}$) and angular rates ($\dot{\theta}, \dot{\phi}, \dot{\psi}$). The fully learned controller outputs directly individual motor velocities ($\Omega$), normalized between -1 and 1. Note that this controller does not require command allocation to the rotors via the allocation matrix $\Delta$, since it outputs motor velocities directly.

\begin{figure}[tbh]
\centering
\includegraphics[width=.8\linewidth]{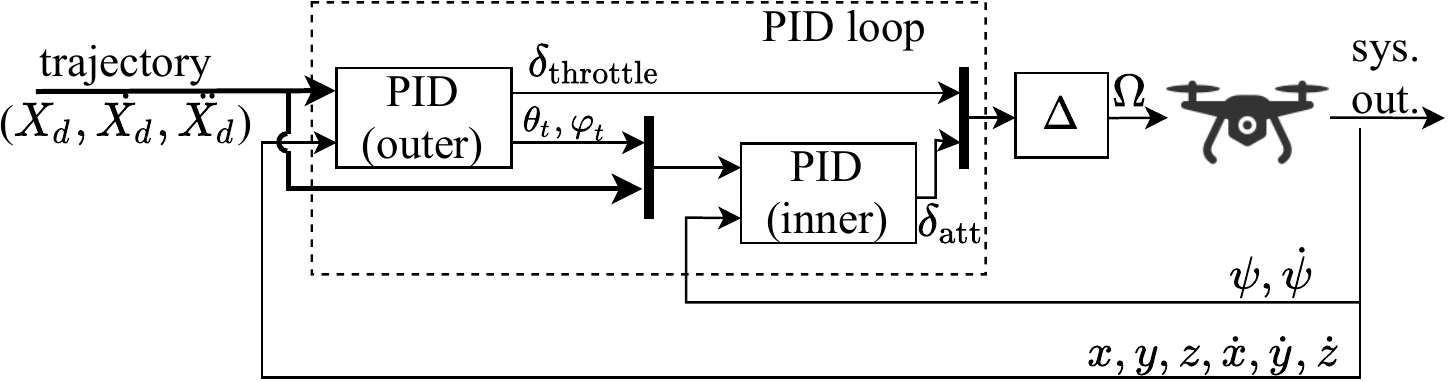}
\caption{Control diagram for the pose PID controller.}
\label{fig:3diag_pid}
\end{figure}

Fig.~\ref{fig:3diag_pid} shows the control loop of two cascade PID controllers, that compose the pose PID controller used in this work. Instead of being guided by the final desired waypoint, as is done with the other controllers, the pose PID tracks a trajectory $X_d(t)$ generated for each waypoint, with parameters, such as duration, maximum velocity and acceleration, predefined via the search algorithm. This pose PID controller consists of two control loops, as is customary for quadrotor controllers. The external loop calculates the pitch and roll targets ($\theta_t, \phi_t$) and throttle force ($\delta_{\textrm{throttle}}$) of the vehicle necessary to track the trajectory. A faster internal loop calculates the lower level attitude commands $\delta_{\textrm{att}}$, needed to track the attitude targets, consisting in roll, pitch and yaw torques. Then, these commands are concatenated with the throttle command, before being allocated to individual motor velocities $\Omega$ via the control allocation matrix $\Delta$.

The trajectory followed by this controller is a straight path from the current ($t_0$) position towards the goal in all 4 directions of movement: $X$, $Y$, $Z$ and yaw. For each direction, the velocity of the trajectory increases linearly from the stationary state towards a predefined maximum velocity, then is maintained constant until it is decreased again towards a stationary state at the goal pose. The parameters of this trajectory are its total duration $T_e$ and the rise time $r_t$, which corresponds to the proportion of $T_e$ which is used to accelerate and decelerate the vehicle. The acceleration needed and maximum velocities in each dimension are automatically calculated based on these constraints. Adequate values for $T_e$ and $r_t$ are found at the same time as the PID parameters through the aforementioned search algorithm and are the same for all directions of movement.

\begin{figure}[tbh]
\centering
\includegraphics[width=\linewidth]{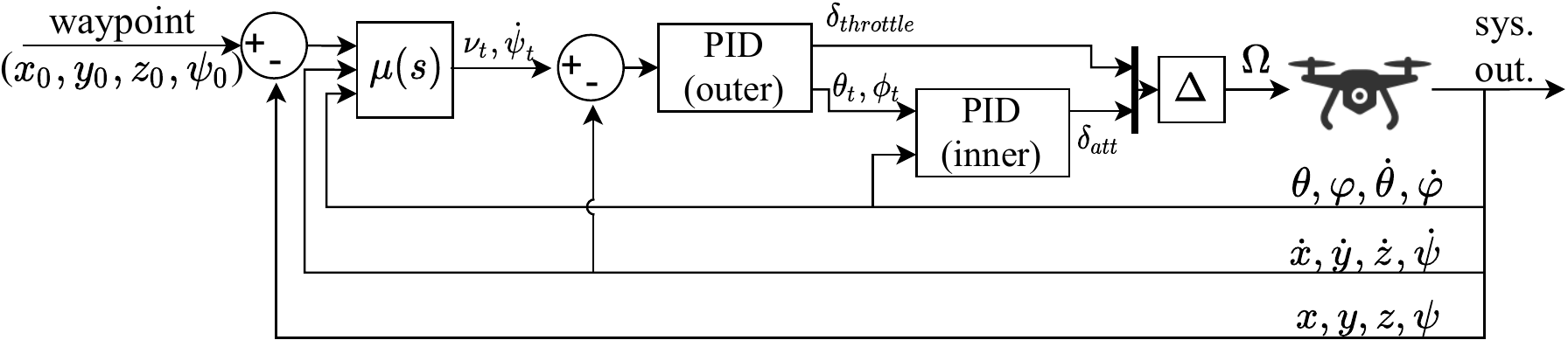}
\caption{Control diagram for the cascade controller: a learned external high-level controller and lower-level PID loops for velocity control, where $\nu_t$ is the linear velocity setpoint and $\dot{\psi}_t$ is the yaw rate setpoint. The $\mu(s)$ block represents the ANN trained using the SAC algorithm.}
\label{fig:3diag_cascade}
\end{figure}

The cascade controller used in this work combines a learned controller $\mu(s)$ and a PID controller, as described by the diagram shown in Fig.~\ref{fig:3diag_cascade}. This controller receives the pose error as input, along with the additional IMU data, in the same way as the fully learned controller presented in Fig.~\ref{fig:3diag_fullrl}.

Attitude setpoints, used as output of the top-level controller of the pose PID stack (Fig.~\ref{fig:3diag_pid}), are usually reserved for aerobatics and "sport settings" of these drones. Instead, existing RL control research opt for "higher" levels of control, such as a discrete displacement step used with DQN-based controllers \cite{wu2018navigating, walvekar2019vision}.
To make use of the continuous action space control capability of actor-critic methods, we opt for a control stack similar to the \textit{Tello} drone, also employed in RL research \cite{model2021belkhale}, which takes, as inputs, the desired 3-dimensional linear velocities ($\dot{x}$, $\dot{y}$ and $\dot{z}$ summarized in the diagram as $\nu_t$) and yaw rate for the quadrotor ($\dot{\psi}_t$), while an underlying internal controller calculates the low level commands needed to navigate towards the given targets.

The aforementioned underlying controller is implemented as a velocity PID controller built with 2 control loops. The inner PID loop returns throttle force and roll, pitch and yaw torques as a command vector $\delta$, allocated to individual motor velocities $\Omega$ via the control allocation matrix $\Delta$. The internal PID controller, similarly to the pose PID controller, has its gains optimized by the CMA-ES search algorithm.

All the proposed controller loops, during experimentation, operate at a 40 Hz sampling rate\footnote{In our model, using a faster PID rate (1000 Hz, for instance) did not lead to significantly better results, but did slow down the simulation significantly.}.

%%%%%%%%%%%%%%%%%%%%%%%%%%%%%%%%%%%%%%%%%%%%%%%%%%%%%
%%%%%%%%%%%%%%%%%%%%%%%%%%%%%%%%%%%%%%%%%%%%%%%%%%%%%
%%%%%%%%%%%%%%%%%%%%%%%%%%%%%%%%%%%%%%%%%%%%%%%%%%%%%
%%%%%%%%%%%%%%%%%%%%%%%%%%%%%%%%%%%%%%%%%%%%%%%%%%%%%

\subsection{PID gains optimization}

Usually, the PID gains for the quadrotor controller are found via manual tuning or via linear model analysis \cite{Park2019}. However, by employing these methods, it becomes increasingly difficult to find gains appropriate for a quadrotor with constant parameters, and even more difficult if the same controller is used for quadrotors with different sets, or varying, parameters, which is the final objective of the controllers described in this section. This occurs because the quadrotor is able to move with six degrees of freedom (DoF), so that, to control movement in every DoF, with 6 uncoupled controllers with proportional, integral and derivative gains each, the final controller would have a total of 18 unique gains to be tuned. A faster and optimized way of tuning PID gains consists of employing an automatic search strategy.

PID gains for both the pose and velocity PID controllers are found using a Covariance Matrix Adaptation (CMA) Evolution Strategy (ES) \cite{cmaes}. This evolutionary search algorithm is a second order approach to estimate a positive definite covariance matrix by an iterative process \cite{cmaes}. This search algorithm is a popular evolutionary approach for optimization problems, shown to be reliable for both local optimization \cite{hansen2001} and global optimization \cite{hansen2004}. Further details on the fitness functions, constraints and parameters used with the \textit{Python} implementation\footnote{$<$\url{https://github.com/CMA-ES/pycma}$>$ Accessed: 29 May 2021} of this search algorithm are laid out in Sect.~\ref{sec:simulations}.

Genetic algorithms have been shown to be efficient for finding optimized PID gains \cite{gapid}. RL could be used also for automatically tuning PID gains \cite{Park2019}. However, developing such methods is beyond the scope of this work.

%%%%%%%%%%%%%%%%%%%%%%%%%%%%%%%%%%%%%%%%%%%%%%%%%%%%%
%%%%%%%%%%%%%%%%%%%%%%%%%%%%%%%%%%%%%%%%%%%%%%%%%%%%%
%%%%%%%%%%%%%%%%%%%%%%%%%%%%%%%%%%%%%%%%%%%%%%%%%%%%%
%%%%%%%%%%%%%%%%%%%%%%%%%%%%%%%%%%%%%%%%%%%%%%%%%%%%%

\subsection{Learning algorithm}\label{sec:learning_algorithm}

As previously mentioned, both learning-based controllers are obtained using the SAC algorithm with automatic reward scale adjustment.

There is a total of five neural networks implemented in this algorithm, which are the two $Q$-networks, for clipped double $Q$-learning, two target $Q$-networks, that track the other two $Q$-networks, and one for the control policy. All the networks are implemented with two hidden layers of 400 and 300 hidden neurons, using a rectified linear unit (ReLU) as activation function, which is a standard architecture for RL algorithms, and is able to cope with a variety of problems using only a single architecture \cite{Hwangbo2017}. The critic networks have no activation function in their outputs and can be evaluated directly as $Q_i(s,a)$. The actor network outputs, however, the parameters for the control policy: $\mu(s)$, with a \texttt{tanh} activation function, and $\ln\sigma(s)$, with a \textit{sigmoid} activation function.

In the actor network, $\mu(s)$ is the mean of the stochastic control policy. The $\ln\sigma(s)$ output is normalized between a pre-set range of $[-9;2]$ and used to add a Gaussian exploration noise $\epsilon$ to $\mu(s)$, with mean 0 and variance equal to the exponential of the normalized output, resulting in the preliminary stochastic policy in:

\begin{equation}\label{eq:3sac_exp_policy}
\pi(s) = \mu(s) + \epsilon,\quad\epsilon\sim\mathcal{N}(0,\sigma(s)).
\end{equation}

The functions $\pi(s)$, $\mu(s)$ and $\sigma(s)$ are then used to estimate the entropy of the stochastic control policy. Finally, $\pi(s)$ is clipped using a \texttt{tanh} function and both it and $\mu(s)$ are scaled to the environment action range, resulting in, respectively, a stochastic and a deterministic control policy usable for training and testing.

In order to simulate the learning process that would take place with an actual quadrotor, the neural network training is done off-line, with a fixed amount of gradient steps, after an episode has ended, regardless of the episode duration, as each update would take too long to be fitted in between live control actions.

\section{Simulations}\label{sec:simulations} % (models, tasks, reward functions, etc.)

Two low-level control tasks performed by the quadrotor are considered: a waypoint guidance task and payload pick up course. The former is mainly used for training the controllers, while the latter is used for evaluation of these controllers, presenting a sudden mass change during payload pick up and drop.

The controllers are trained and tested in a simulated environment, consisting of the dynamic model of a quadrotor, without accounting for aerodynamic effects and sensor disturbances. This dynamic model takes as input the desired angular velocities of each rotor, while providing, as observations, the linear and angular velocities of the quadrotor in the local frame, and the linear and angular positions in the setpoint frame. Additionally, the dynamic model allows, at any time, the adjustment of five key quadrotor parameters: propeller diameter, arm length, hub size and vehicle mass.

During training, the quadrotor simulation is subject to motor command saturation. For testing, besides the motor saturation and the sudden change in mass of the payload pick up course, the quadrotor is subject to additional sensor noise and motor command delay.

The deep learning models developed are processed at the institution's remote GPU cluster, with 4$\times$ \textit{Nvidia GTX 1080Ti} 12 GB VRAM GPUs, 2$\times$ \textit{Intel Xeon E5-2669 v3} CPUs and 128 GB RAM.

%%%%%%%%%%%%%%%%%%%%%%%%%%%%%%%%%%%%%%%%%%%%%%%%%%%%%%%%%%%%%%%%%%%%%%%%%%%%%%%%%%%%%%%

\subsection{Training environment}\label{sec:llc-train}

The task for the quadrotor during training is to navigate towards the origin, starting from a random point. For different training episodes, the environment parameters are randomized, following the bounds and constraints described by Table~\ref{tab:4parameter_table}. The table includes, as well, the description of the fixed environment parameters for the benchmark learned controller. Arm length is constrained with respect to the propeller diameter in order to properly accommodate it, while vehicle mass is constrained with respect to the same parameter in order to maintain a reasonable thrust-weight ratio, particularly around a mean rotor velocity of 5323 RPM, which was found to be the hover point of the quadrotor with fixed parameters described in Table~\ref{tab:4parameter_table}.

\begin{table}[h]
\centering
\caption{Upper and lower bounds of randomized quadrotor parameters used for waypoint controller development and PID gains optimization. Fixed parameters are used for training the single-environment benchmark controller.}
\label{tab:4parameter_table}
\begin{tabular}{@{}llll@{}}
\hline\noalign{\smallskip}
Parameter                     & Lower bound     & Upper bound      & Fixed values \\ \noalign{\smallskip}\hline\noalign{\smallskip}
Propeller diameter ($d$) (in) & 6.0             & 12.0             & 10.0 \\
Hub size (m)                  & 0.05            & 0.15             & 0.10 \\
Arm length (m)                & $0.0167d + 0.05$ & $0.0334d + 0.05$& 0.3 \\
Vehicle mass (kg)             & $0.1d - 0.3$    & $0.265d - 0.9$   & 1.2 \\ \noalign{\smallskip}\hline
\end{tabular}
\end{table}

The training hyperparameters used for the learned controllers are described by Table~\ref{tab:3hyper_table}. Optimizer, replay buffer size, initial training steps and target update interval were chosen based on existing literature. The other hyperparameters were found by manual tuning, aiming to strike a balance between steady state guidance error and computational time, tested in ranges approximately one order of magnitude ($\times10$) below and above the values described by the same table. In our implementation, different target entropies ($h$) did not significantly affect the learning performance, therefore this parameter is set at $-4$, as suggested by the original author of the algorithm for 4-dimensional action spaces \cite{haarnoja2019soft}.

\begin{table}[h]
\centering
\caption{Hyperparameter and experiment configuration for the control tests.}
\label{tab:3hyper_table}
\begin{tabular}{@{}ll@{}}
\hline\noalign{\smallskip}
Parameter                                  & Value                                             \\ \noalign{\smallskip}\hline\noalign{\smallskip}
Optimizer                                  & Adam                        \\
Learning rate (Actor)                      & $3\cdot10^{-5}$                                   \\
Learning rate (Critic)                     & $3\cdot10^{-4}$                                   \\
Discount ($\gamma$)                        & $0.99$                                            \\
Replay buffer size                         & $10^7$                                            \\
Initial training steps                     & $10^4$                                            \\
Number of samples per minibatch            & $256$                                             \\
Entropy target ($h$)                       & $-4$                      \\
Target smoothing coefficient ($\tau$)      & $0.005$                                           \\
Target update interval                     & $1$                                               \\
Gradient steps per episode                 & $128$                                             \\ \noalign{\smallskip}\hline
\end{tabular}
\end{table}

The ANNs are structured as described in Sect.~\ref{sec:methodology}, where the critic networks receive, as input, the state-action pair, for which they output its predicted value. The actor network receives, as input, the state for which it predicts an appropriate action. As output, the network provides the deterministic action prediction and the natural logarithm of the appropriate exploration rate, with which a stochastic control policy can be computed.

\subsubsection{PID gains optimization}

Here, 18 different PID gains are considered for a quadrotor controller, as well as 2 trajectory parameters for the tracking controller. These gains are classified in categories with respect to which variable they control (linear space: $x,y,z$, angular space: $\theta,\phi,\psi$) and which term they refer to (proportional: \textbf{p}, integral: \textbf{i}, derivative: \textbf{d}). For example, the pitch angle proportional controller is denoted as $k_{\textrm{p},\phi}$.

Prior to initiating the CMA-ES search algorithm, 100 evaluation environments are generated using randomized parameters within the bounds and constraints described in Table~\ref{tab:4parameter_table}. This is done to maintain consistency between evaluations of different sets of PID gains. The search space is constrained using a simple sinusoidal function normalized between lower and upper parameter boundaries.
For the pose PID controller, the lower bounds are $k_{\textrm{p},x},k_{\textrm{p},y}$:0.1, $k_{\textrm{p},z}$:0.5, $k_{\textrm{p},\theta},k_{\textrm{p},\phi}$:1.0, $k_{\textrm{p},\psi}$:0.5 and 0 for all other gains. The upper bound is set to 10 for all gains. The trajectory parameters $T_e$ and $r_t$ have, respectively, lower bounds of 1.0 and 0, and upper bounds of 10.0 and 0.5.
The velocity PID controller shares the same gain boundaries as the pose controller, except for the lower bound of the $k_{\textrm{p},\psi}$ gain, which is increased to 5.0.
Lower bounds were initially chosen as 0 for all gains. However, the search would result in 0 for some of the gains, including some proportional gains and, sometimes, eliminating completely the control feedback loop. This happened especially with the proportional yaw gain ($k_{\textrm{p},\psi}$), due to its uncoupled nature not affecting overall stability. Because of this, the lower bounds were raised to adequate values for the indispensable proportional gains. Upper bounds were chosen arbitrarily as 10. Since gains were not clipped at 10 during search, it was left as is.

The CMA-ES search is performed for 1000 iterations with initial standard deviation $\sigma$ of 2.0. In the training runs performed, different initial search vectors yield different results. Thus, the initial vectors that result in the best set of gains for each PID controller are used. For the velocity PID gains, the initial search vector has a value of 5.0 for all entries. For the pose PID gains, however, the initial search vector is obtained from manual tuning, set to values that allow the controller to complete the waypoint guidance task successfully at least once.

\paragraph{Evaluation - position controller:} The performance of a set of controller gains in each iteration is evaluated as the average sum of position and attitude errors, with respect to the final goal, of each evaluation environment. If the quadrotor crosses the environment boundary or destabilizes, the performance score is 80, instead. Thus, the evaluation score of each run corresponds to minus the reward function described further in Eq.~\ref{eq:4ll_rew_function}. Measuring errors with an inertial reference, instead of using the trajectory, takes into account the time taken to reach the goal waypoint, being preferred for this kind of task. For consistency, this waypoint is always set to the origin, with a starting point set to a fixed distance of 1 m, with initial quadrotor attitude angles ranging from -0.064$\pi$ to 0.064$\pi$ radians (approximately 11.5°) for pitch and roll, which are reasonable ranges for non-acrobatic quadrotors, and from -0.3$\pi$ to 0.3$\pi$ radians (approximately 54°) for yaw, which is a reasonable range to account any need for turning left or right. Starting points are randomized in advance and are the same for every iteration of the search algorithm.

\paragraph{Evaluation - velocity controller:} Starting from the origin, the quadrotor follows a predefined trajectory, acting in all four action dimensions: roll, pitch, yaw and throttle. This trajectory consists of 5 seconds pulses of 1.0 $m/s$ for the setpoints of $\dot{x}$, $\dot{y}$ and $\dot{z}$, and a pulse of 1.0 $rad/s$ for the setpoint of $\dot{\psi}$, one pulse at each time. Then, these pulses are applied again with negative velocity. The performance is evaluated as the sum of the velocity error along this trajectory for all 4 degrees of freedom. This error ($e$) is soft-clipped at the value of 200 to reduce the influence of unstable environments in the evaluation score for each candidate, thus being expressed by:

\begin{equation}\label{eq:4cmaes_softclip}
e_{clip} =
  \begin{cases}
    e & \text{if $e \leq 200$,}\\
    200 + \sqrt{e-200} & \text{if $e > 200$.}
  \end{cases}
\end{equation}

\noindent This way, the search algorithm prioritizes a set of gains that works very well, i.e. with minimal error, with a narrower set of environments, than gains that display more errors, but are applicable to a broader set of environments.

\paragraph{Search dimensionality reduction:} Taking advantage of the symmetry of the quadrotor to further speed up the search time, the gains for pitch and roll angles are the same, as well as gains for the movement along the X and Y axes. With these optimizations, the search space is effectively reduced to only 12 parameters: $k_{\textrm{p},xy}$, $k_{\textrm{p},z}$, $k_{\textrm{i},xy}$, $k_{\textrm{i},z}$, $k_{\textrm{d},xy}$, $k_{\textrm{d},z}$, $k_{\textrm{p},\theta\phi}$, $k_{\textrm{p},\psi}$, $k_{\textrm{i},\theta\phi}$, $k_{\textrm{i},\psi}$, $k_{\textrm{d},\theta\phi}$ and $k_{\textrm{d},\psi}$, with 2 extra parameters, $T_e$ and $r_t$, for the pose PID gains optimization.

\subsubsection{Reinforcement learning environment settings}

For the training phase, the quadrotor is subject to a 20 Hz control rate and a maximum episode time of 60 seconds (1200 time steps). For testing and experimentation, the same controller is subject to a 40 Hz control rate and maximum episode time of 60 seconds (2400 time steps). This is done to prevent a jittering behaviour we found during experiments, that occurs if the control rate is too low, likely caused by overfitting in some of the models trained.

The learning-based controller takes as input the pose error of the quadrotor with respect to the setpoint in terms of 3-dimensional errors ($x,y,z$) and yaw ($\psi$), as well as IMU data, such as roll, pitch, attitude rates and linear velocities. As output, the controller yields a normalized motor velocity vector. Due to the variability of vehicle parameters, the action is not always centered around the hover point. Consequently, the controller must learn this offset, an especially useful ability for handling the picking up and dropping of the payload.

The equations of motion in the local and inertial frames are, then, evaluated in the discrete-time environment using the ODE integrator included in \textit{Python}’s \textit{scipy}\footnote{$<$\url{https://www.scipy.org/}$>$ Accessed: 12 Mar. 2021} library, employing the Real-valued Variable-coefficient Ordinary Differential Equation solver (VODE) and a method based on backward differentiation formulas (BDF). The system is integrated in up to 500 steps per control step, but sampled at the desired control rate for training or testing.

For the origin guidance task, for both training and testing, the starting points are chosen in the same fashion as in the PID gains search, described in the paragraph "\textbf{Evaluation - position controller}", but sampled randomly at every episode, instead of being chosen from a set of predefined parameters.

To complete this task, we define a reward function which is densely distributed through states, unless a state is terminal, in other words, if the quadrotor crosses the environment boundaries, in which then a reward of -80 is given. The reward function thus is described by:

\begin{equation}\label{eq:4ll_rew_function}
r(s,a,s') =
\begin{cases}
  -80 & \hspace{-5.5em} \textrm{if the state is terminal;} \\
  -1.0 \sqrt{ p_x^2 + p_y^2 + p_z^2 } -0.1 \sqrt{ \theta^2 + \phi^2 } -0.5 \sqrt{ \psi^2 } & \textrm{otherwise,}
\end{cases}
\end{equation}

\noindent based on the position errors ($p_x, p_y, p_z$) and the attitude angles: roll ($\theta$), pitch ($\phi$) and yaw error ($\psi$). Defining the reward function in terms of the inverted quadratic error is customary in control problems \cite{Engel2014}. However, we found in preliminary experiments that, for this environment, the root-squared error produces a more accurate control policy, when compared to the quadratic error. The reward term for roll and pitch angles aims to improve stability by minimizing the pitch and roll of the quadrotor during training. The weight factors for the positions and angles terms were found by manual tuning, where we did not want the angle terms to dominate the final reward too much.

%%%%%%%%%%%%%%%%%%%%%%%%%%%%%%%%%%%%%%%%%%%%%%%%%%%%%%%%%%%%%%%%%%%%%%%%%%%%%%%%%%%%%%%

\subsection{Testing environment and experiment setup}\label{sec:llc-test}

The evaluation of the proposed controllers take place in a payload pick up course. This course consists in two waypoints in an enclosed 3D space, one near the bottom of this space and another near the top, towards which the quadrotor navigates using the existing waypoint guidance controllers. The quadrotor starts the course near the lower waypoint, which becomes its first target. Upon reaching this target, the vehicle "picks up" a simulated payload that needs to be delivered to the upper waypoint, where it will be released. Finally, to finalize the course, the quadrotor returns empty to the bottom waypoint. This task is designed to take advantage of the parameter-agnostic capability of the controller by changing the vehicle size and mass mid-flight and requiring this mass to be carried upwards.

The payload pick up act is emulated by increasing the quadrotor mass by 30\% of its current mass, which is a reasonable payload-mass ratio considering real life payload-carrying multi-rotor (non-winged) drones, and by increasing its hub size by an extra 10 centimeters radius.

The quadrotor is considered to have reached a waypoint when it is at most 0.15 m from its destination, which is set considering the steady state errors of the controllers, and is considered stable enough for safely picking up a payload. For the quadrotor to be considered stable, the root mean squared errors of its angular positions and rates must be below a threshold which is defined by manual tuning, in this case, 0.15, where angular positions are measured in radians and angular rates are measured in $rad/s$.

While RL controllers are trained using starting points 1 meter apart at most from the waypoint target, the payload pick up course presents waypoints apart a greater distance. Operating outside these boundaries used for training can result in unpredictable behavior for ANN-based controllers. Thus, if the vector that points from the quadrotor towards the target waypoint is grater than 1 meter, it is clipped to this maximum length and the controller attempts to reach this clipped target instead. Since the pose PID controller is guided by a trajectory instead of waypoints, it does not need such adaptation

For ease of visualization and evaluation, the experiments are performed in environments varying with respect to only propeller diameter and vehicle mass, with the arm length adjusted accordingly to a mean between the upper and lower bounds set for the training phase, with no randomization. This way, the controller performance can be evaluated in a 2-dimensional visualization. The ranges and constraints of these three parameters are described in Table~\ref{tab:4test_ranges}.

\begin{table}[h]
\centering
\caption{Range of quadrotor parameters used in experiments.}
\label{tab:4test_ranges}
\begin{tabular}{@{}llll@{}}
\hline\noalign{\smallskip}
Parameter                     & Lower bound      & \begin{tabular}[c]{@{}l@{}}Upper bound\\ (waypoint tracking)\end{tabular} & \begin{tabular}[c]{@{}l@{}}Upper bound\\ (payload pick up)\end{tabular} \\ \noalign{\smallskip}\hline\noalign{\smallskip}
\begin{tabular}[c]{@{}l@{}}Propeller diameter\\ ($d$) (in)\end{tabular}
& 6.0              & 12.0             & 12.0  \\
Arm length (m)                & $0.025d + 0.05$       & $0.025d + 0.05$       & $0.025d + 0.05$  \\
Vehicle mass (kg)             & $0.1d - 0.3$     & $0.265d - 0.9$   & $0.2d - 0.66$  \\ \noalign{\smallskip}\hline
\end{tabular}
\end{table}

For the payload pick up task, the vehicle mass upper bound is reduced to approximately 76\% of the value used for the training task, in order to accommodate the extra 30\% payload weight, without burdening the vehicle to a point it cannot hover. When the extra mass is added, the total mass equals the upper bound set for the training task.

The waypoints for the payload pick up task are the same for all experiments for consistency. The pickup point is $[-0.8,-0.3,-0.5]m$, in the lower half of the environment, whereas the drop point is $[0.8,0.3,0.5]m$, in the upper half, requiring the quadrotor to climb while carrying the extra weight, posing an interesting challenge for the designed controllers.

The quadrotor spawns close to the pickup point, anywhere within a 0.5 m cube centered on it and with the same initial attitude angles used in training, which range from -0.064$\pi$ to 0.064$\pi$ radians for pitch and roll and from -0.3$\pi$ to 0.3$\pi$ radians for yaw.

For each controller, enough tests are performed until it reaches 100 successful samples. A payload pick up test is considered successful if the quadrotor manages to reach all the waypoints defined in the trajectory within a time limit, set to 500 time steps (12.5 seconds at 40 Hz) for each waypoint.

The disturbances included for testing and experimentation simulations are a random Gaussian noise added across all the sensor readings with mean 0 and standard deviation of 0.05, and a motor command delay using a low pass exponential filter described by:

\begin{equation}\label{eq:3sac_exp_policy}
\bar{\Omega}_t = 0.5 \Omega_t + 0.5 \Omega_{t-1},
\end{equation}

\noindent which aims to emulate the internal motor dynamics.

\section{Results and discussion}\label{sec:experiments}

For the waypoint guidance experiments, starting conditions are the same as used during training, described in Sect.~\ref{sec:llc-train}. Success and failure conditions are also the same: in a successful test, the quadrotor is capable of remaining within the environment boundaries for 12.5 seconds. If these boundaries are crossed before this time, the experiment is considered a failure. Note that this does not take into account the quadrotor stability, errors and oscillatory responses.

For the payload pick up experiments, starting conditions are also the same used during training, described in Sect.~\ref{sec:llc-train}, except for the fact that initial states are resampled. Successful experiments consist of the quadrotor reaching the pickup point again after dropping the payload in the designed drop point. If the quadrotor crosses the environment boundaries, or is unable to reach any waypoint within 12.5 seconds from spawning or from the previous waypoint, the experiment is considered a failure. Further details on the criterions used for convergence to a waypoint can be found under Sect.~\ref{sec:llc-test}.

One of the metrics used to evaluate the performance of each controller in carrying out the proposed tasks is its expectation ($\mathbb{E}[x]$) to do so, which is estimated for different combinations of the quadrotors size (propeller diameter and arm length) and mass. The experiments for each controller are distributed in a 2D parameter space, composed by the propeller diameter and the vehicle mass, according to the parameter combination used for each test. Then, the expectation map is computed using a moving window filter of size $[1\times0.417]$ (propeller size$\times$mass) in a 100$\times$100 grid in the parameter space. For each point in the map, the expectation of success is estimated by the amount of successful tests under a window around this point, with respect to the total tests under this window, or $n_\textrm{success}/n_\textrm{total}$. Points outside the limits established by Table~\ref{tab:4test_ranges} and points in which there are no tests under its respective window are left blank.

For each RL-based controller, 10 training runs are performed to demonstrate the stability and robustness of the training process. Out of these training runs, the one that results in the "best" controller, according to a success criteria, is chosen for subsequent experiments, which consists in choosing the controller which attains the highest success rate specifically in the payload pick up task, which is the final goal of these controllers. The performance of the chosen controller is evaluated over 100 tests using a separate set of 100 quadrotor parameter samples.

\subsection{PID tuning and performance}

The resulting gains for the controllers after the search is concluded are described by Table~\ref{tab:5gain_table_p} for the pose PID gains and by Table~\ref{tab:5gain_table_v} for the velocity PID gains. Additionally, the search algorithm found a trajectory for the pose PID with 10.0 seconds total duration, with rise time equivalent to 33.5\% of the trajectory time.

\begin{table}[tbp]
\centering
\caption{PID gains for the quadrotor position control.}
\label{tab:5gain_table_p}
\begin{tabular}{@{}llll@{}}
\hline\noalign{\smallskip}
                            & P          & I          & D          \\ \noalign{\smallskip}\hline\noalign{\smallskip}
Longitudinal velocity (X,Y) & 0.52       & 0.00       & 0.40        \\
Vertical velocity (Z)       & 7.28       & 8.24       & 0.69        \\
Roll ($\theta$), Pitch ($\phi$) & 10.00   & 0.01      & 3.15       \\
Yaw ($\psi$)                & 3.62      & 1.71        & 4.09        \\ \noalign{\smallskip}\hline
\end{tabular}
\end{table}

\begin{table}[tbp]
\centering
\caption{PID gains for the quadrotor velocity control.}
\label{tab:5gain_table_v}
\begin{tabular}{@{}llll@{}}
\hline\noalign{\smallskip}
                            & P          & I          & D          \\ \noalign{\smallskip}\hline\noalign{\smallskip}
Longitudinal velocity (X,Y) & 0.14       & 0.00       & 0.00       \\
Vertical velocity (Z)       & 4.96       & 1.54       & 0.00       \\
Roll ($\theta$), Pitch ($\phi$) & 9.88   & 0.51       & 0.96       \\
Yaw ($\psi$)                & 9.78       & 0.00       & 3.41        \\ \noalign{\smallskip}\hline
\end{tabular}
\end{table}

\begin{figure}[tbp]
    \centering
    \begin{subfigure}[t]{.33\linewidth}
        \centering
        \includegraphics[width=\linewidth]{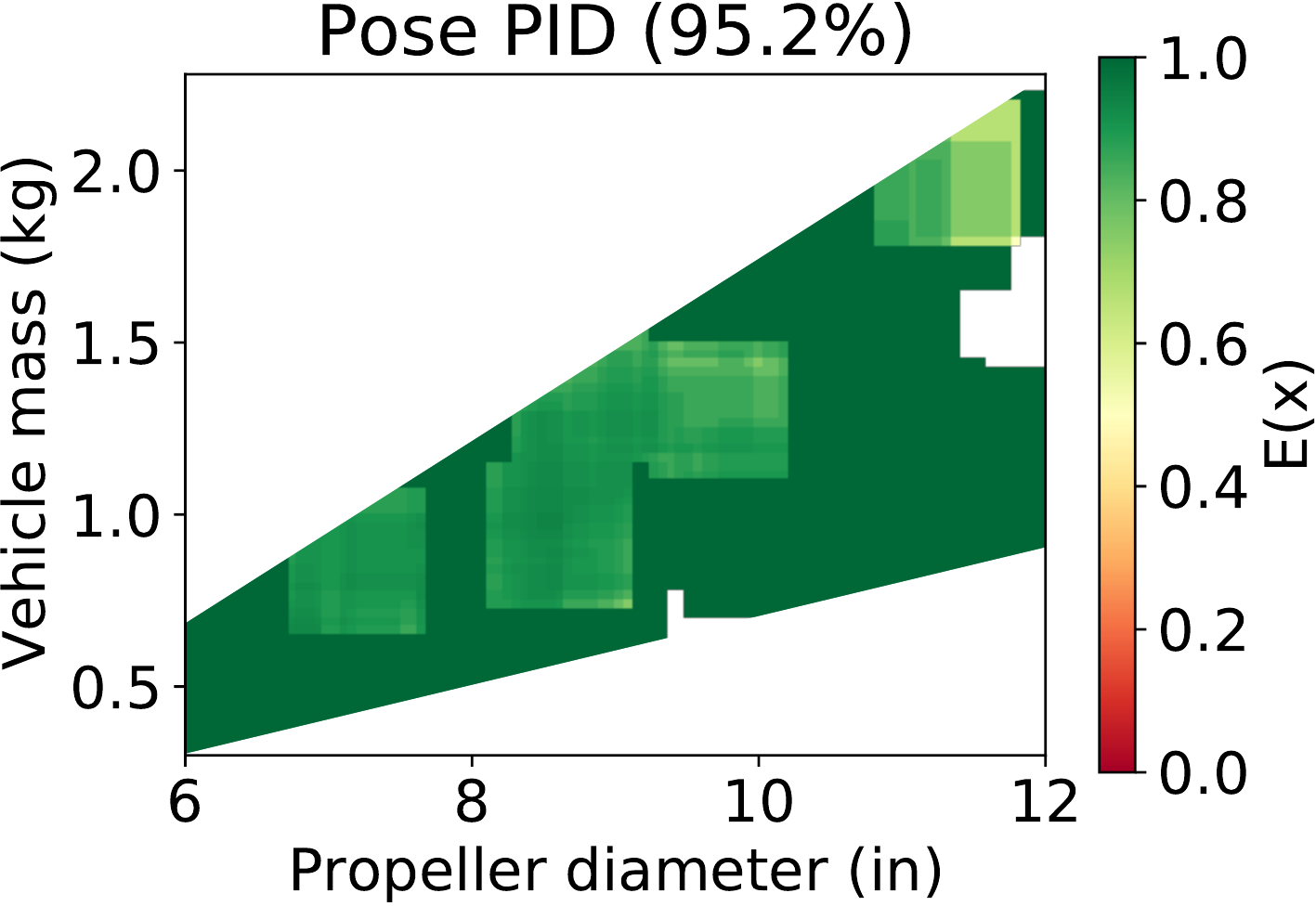}
        \caption{}
    \end{subfigure}%
    ~
    \begin{subfigure}[t]{.33\linewidth}
        \centering
        \includegraphics[width=\linewidth]{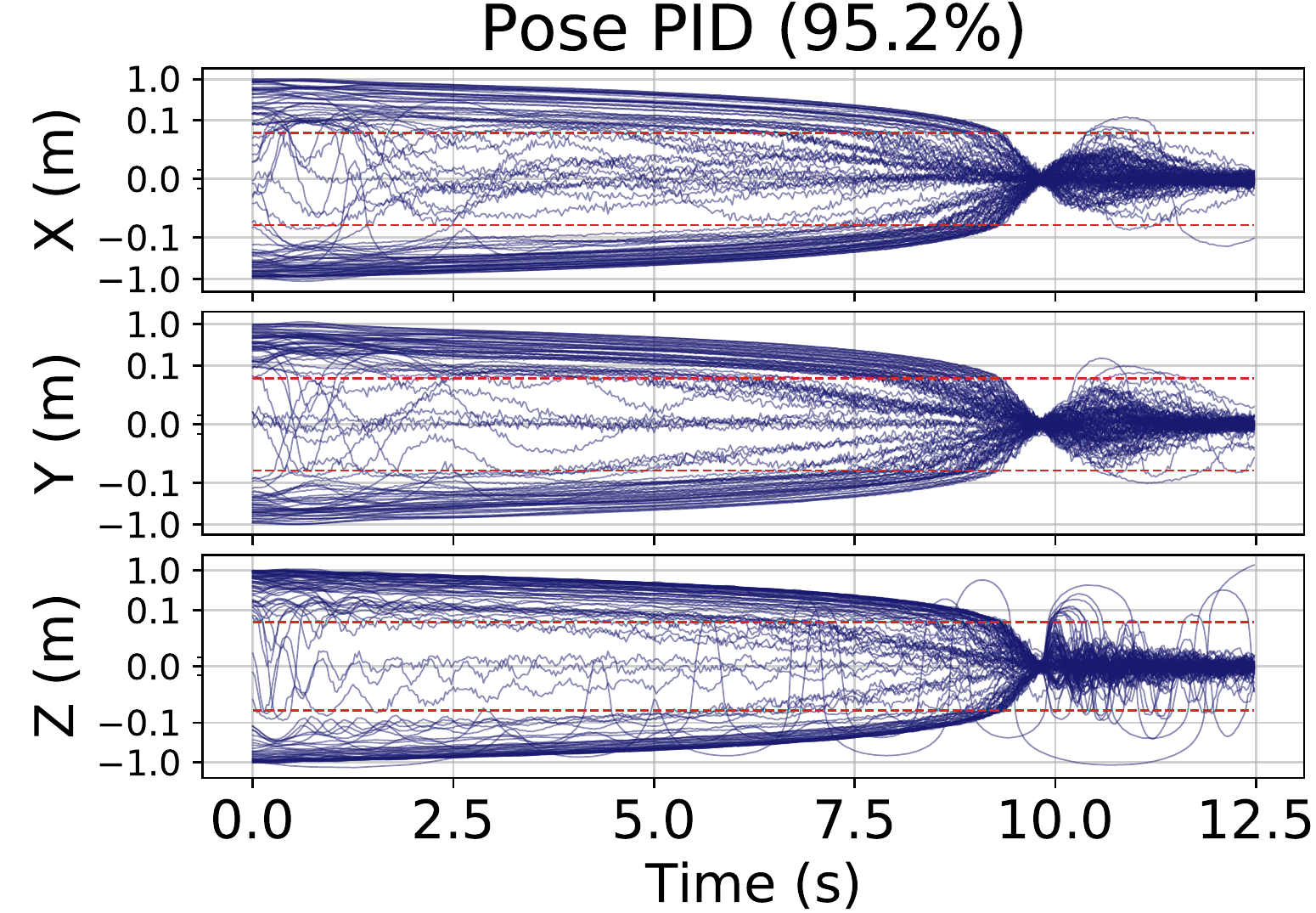}
        \caption{}
    \end{subfigure}%
    ~
    \begin{subfigure}[t]{.33\linewidth}
        \centering
        \includegraphics[width=\linewidth]{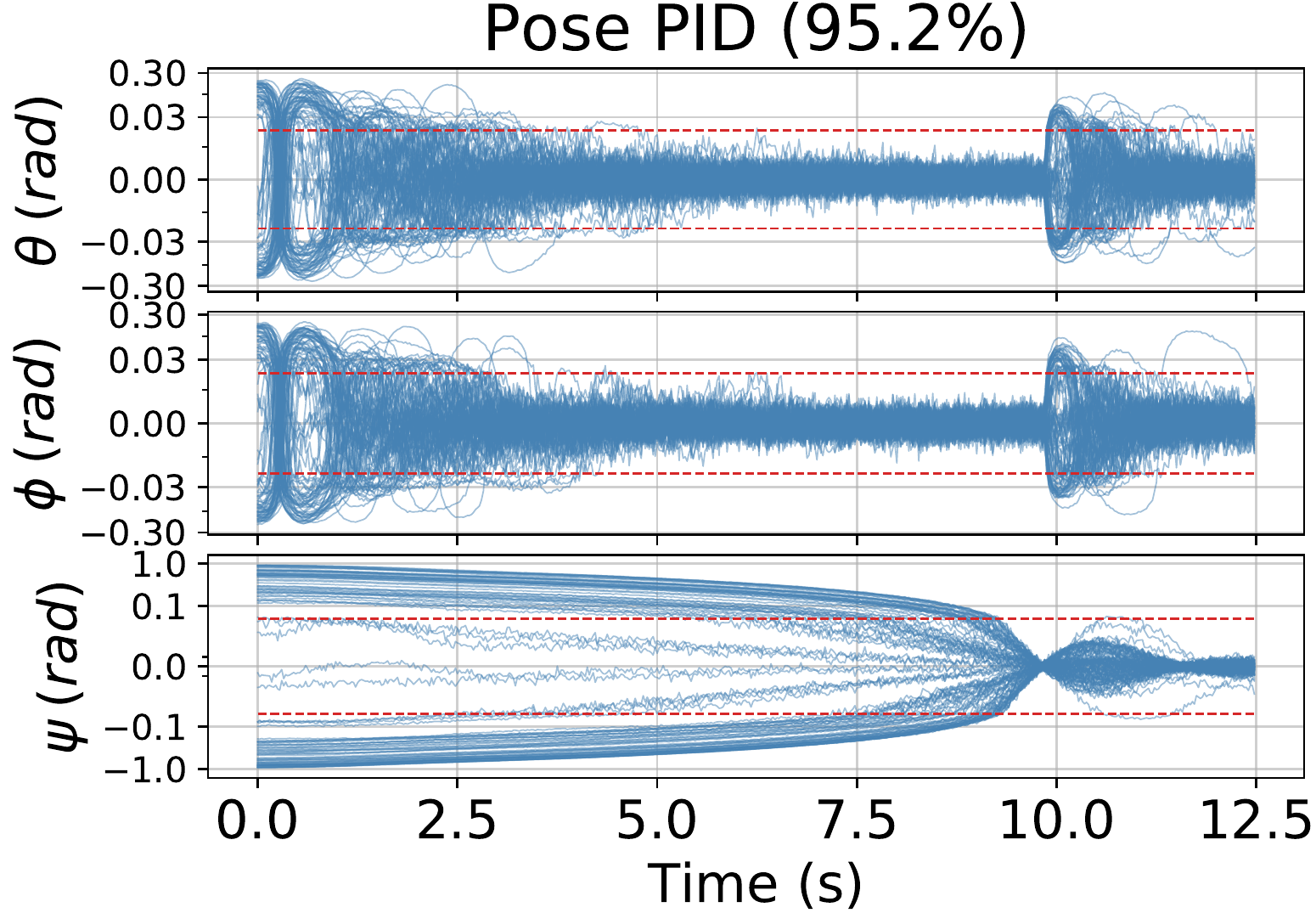}
        \caption{}
    \end{subfigure}

    \caption{Statistics for the pose PID controller. (a) Expected success rate in the waypoint guidance task. (b) Position of the quadrotor over time. (c) Attitude of the quadrotor over time.}
    \label{fig:5results_pid}
\end{figure}

Fig.~\ref{fig:5results_pid} shows the resulting performance of these gains for the pose PID in the waypoint guidance task. This controller obtained 100 successful tests out of 105. Fig.~\ref{fig:5results_pid}b shows the position over time of all successful experiments broken down in each direction of movement, X, Y and Z. Note that the horizontal and vertical movements behave differently. Fig.~\ref{fig:5results_pid}c shows the attitude over time of all successful experiments broken down in each direction of rotation, $\theta$, $\phi$ and $\psi$.

The position axes have been symmetrically log-scaled\footnote{The plot is $\log$-scaled for most of the plot area, except a small region around the axis origin, where the plot is linearly scaled. This scale is appropriate for data points with negative values in the scaled axis.} so that the quadrotor behaviour around the origin is highlighted. The red dashed line represents the extent of the linear scaling area, which is 0.05 m to each direction around the origin, where 95\% of the initial distance has been covered. This visualization, though unrefined, contains abundant information about the behaviour of the controller.

A preliminary highlight about the trajectories generated by the quadrotor while controlled by the pose PID controller, displayed in Figs.~\ref{fig:5results_pid}b and \ref{fig:5results_pid}c, is how it closely follows the predefined setpoint trajectory with 10 seconds duration, overshooting the target destination after this time has passed, while the setpoints remain at the origin for both position and velocity, producing the "bottleneck" effect observable in the figures. Albeit, at first glance, the trajectory settings found by the CMA-ES search seem to be too slow and conservative, decreasing the trajectory time and increasing the maximum velocity of the quadrotor abruptly decreases the success rate of this controller for multiple environments. These results are further discussed in Sect.~\ref{sec:5wpt_results}, in direct comparison to the results obtained from the main learning-based controllers.

%%%%

\subsection{Single environment training and performance}

Despite the learning converging quickly to a satisfactory control policy with low steady state errors, training the waypoint guidance controller with fixed environment parameters put this controller at risk of overfitting to this particular environment and to its lack of disturbances. Without the randomization of training environment parameters, 10 training runs were performed, but none were able to complete the payload pick up course with the additional disturbances applied during testing. Without the disturbances, however, 9 out of the 10 resulting controllers could complete the course.

\subsection{Performance evaluation of the waypoint guidance task} \label{sec:5wpt_results}

\begin{figure}[tbp]
\centering
\includegraphics[width=.75\linewidth]{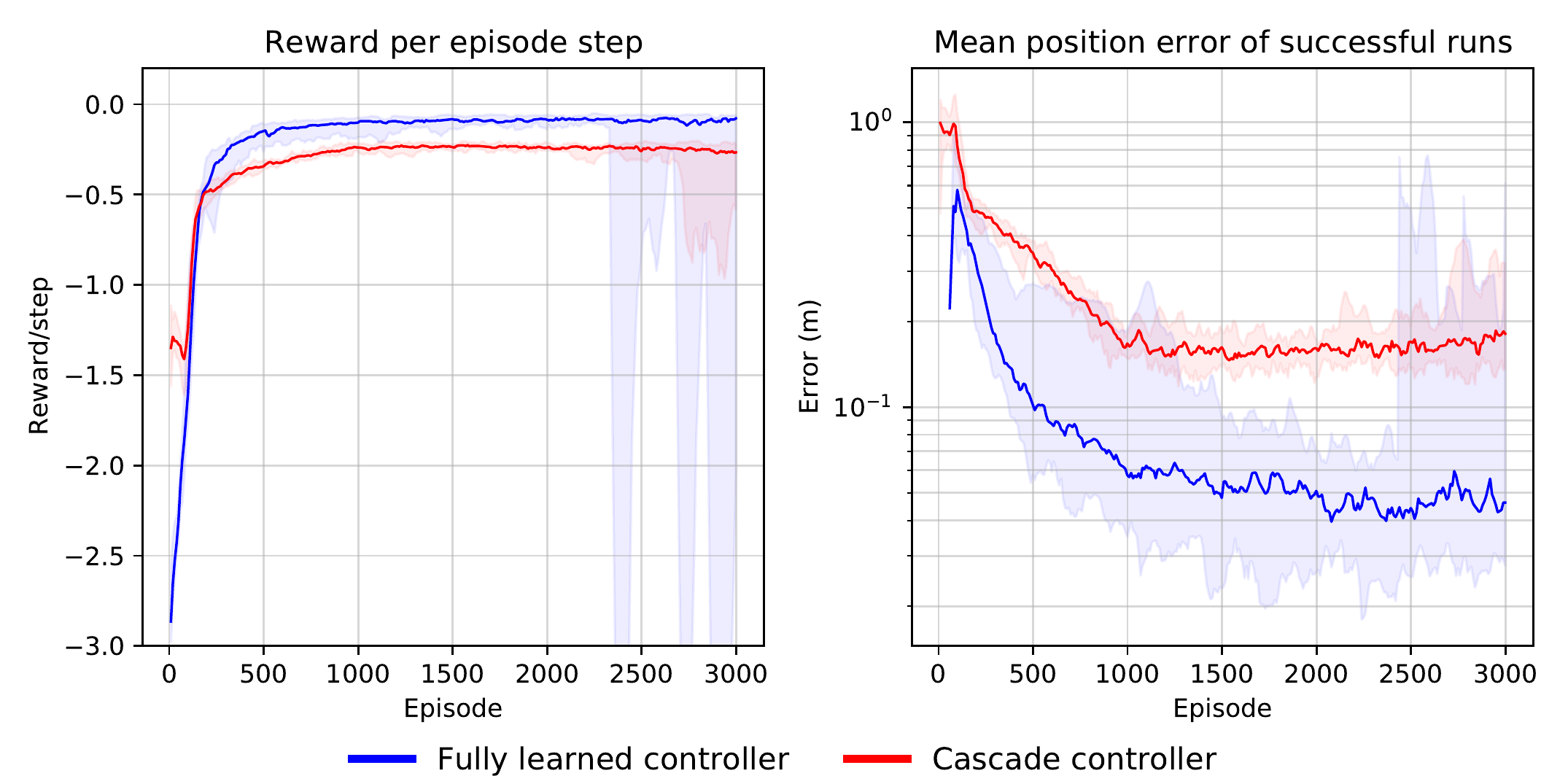}
\caption{Training metrics for the learned controllers. The lines refer to the median metric from the 10 training runs, while the shaded region denotes the minimum and maximum metrics from these runs. Each data point represents the moving average of the 100 previous tests (50 previous episodes).}
\label{fig:5training_curves}
\end{figure}

Fig.~\ref{fig:5training_curves} shows the training performance of 10 different training runs of 3000 episodes for each of the main learned controllers: the fully learned and cascade controllers. In the graph, the bold lines refer to the median metric from the 10 training runs, while the borders of the shaded region denotes the minimum and maximum metrics from each run.

Both the fully learned controller and the cascade controller take about the same amount of training episodes to converge to a locally optimal policy, although displaying very disparate position errors, with the fully learned controller performing significantly better than the cascade controller.

For the fully learned controller, 9 out of 10 training runs result in usable controllers, using weights obtained after 3000 episodes. The single failed training run is the probable cause of the divergent metrics observable after episode 2000 in the graph. On the other hand, all of the cascade controllers resulting from their training runs are able to carry out the proposed tasks. The training runs selected according to the success rate criteria display performance close to the median training runs, which also displayed good performance, within the margins shown in Table~\ref{tab:5curves_statistics}, for both controllers, further confirming that the training process satisfactorily is robust.

\begin{figure*}[tbp]
\centering
\includegraphics[width=\linewidth]{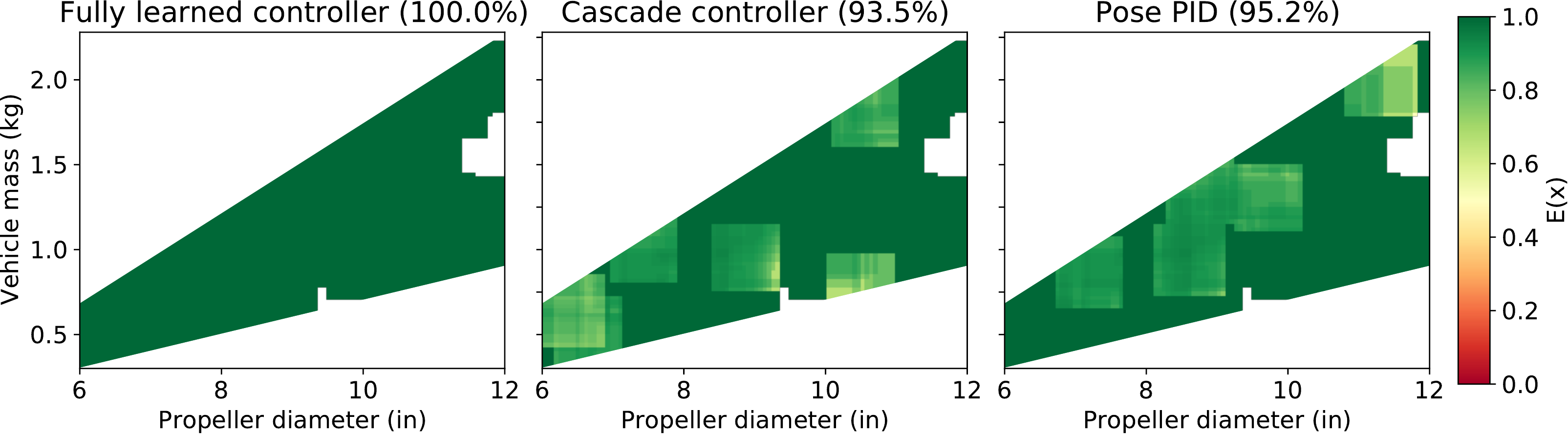}
\caption{Expected success rate in the waypoint guidance task for each controller by combination of quadrotor parameters.}
\label{fig:5filter_origin}
\end{figure*}

Fig.~\ref{fig:5filter_origin} shows the expectation of success in completing the waypoint guidance task for the learned controllers and the baseline pose PID controller.

The fully learned controller performs better overall in the waypoint guidance task, with a slightly higher robustness to a variety of environment parameters than the controllers that rely on an internal PID loop, even though the gains for these controllers were optimized for the displayed range of quadrotor parameters. The pose PID, however, still performs slightly better than the cascade controller.

\begin{figure*}[tbp]
\centering
\includegraphics[width=\linewidth]{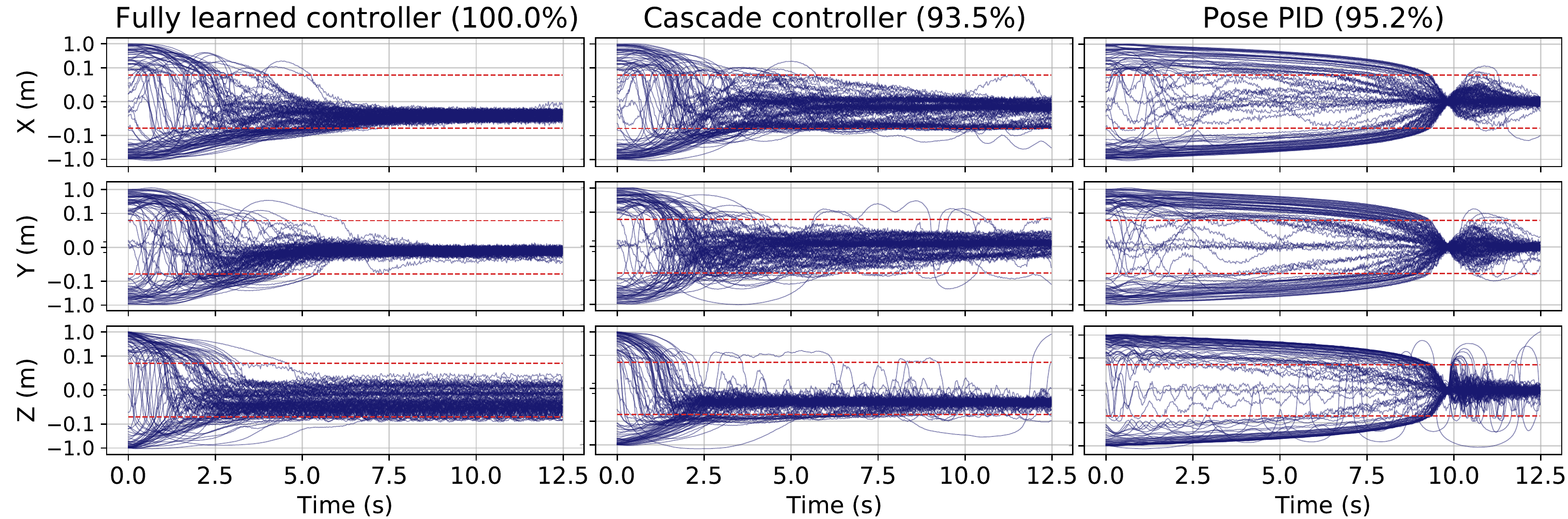}
\caption{Position of the quadrotor over time for different controllers in the waypoint guidance task.}
\label{fig:5origin_curves}
\end{figure*}

\begin{figure*}[tbp]
\centering
\includegraphics[width=\linewidth]{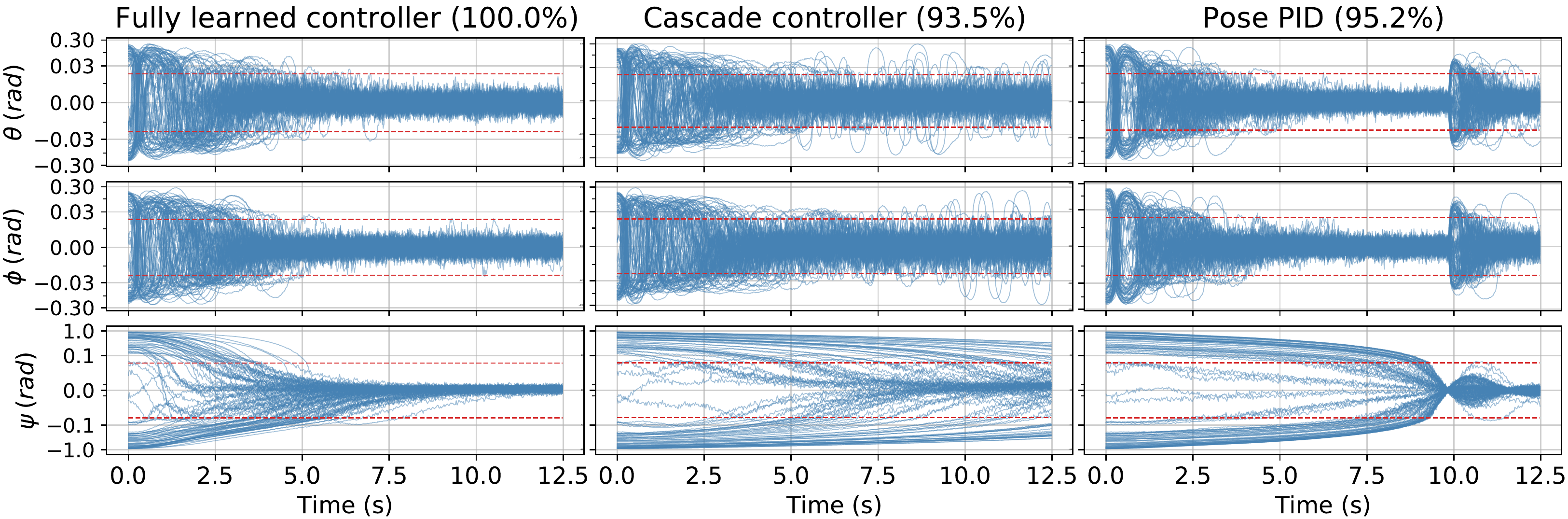}
\caption{Attitude of the quadrotor over time for different controllers in the waypoint guidance task.}
\label{fig:5origin_curves_angles}
\end{figure*}

As a further inquiry into this task, for all 100 successful tests for each controller, over time, Fig.~\ref{fig:5origin_curves} shows the position of these tests and Fig.~\ref{fig:5origin_curves_angles} shows the attitude.

The first highlight would be the settling time of the quadrotor. The fully learned controller has a much faster approach to the origin compared to the other controllers, converging, in average, in 6.21 seconds, while the other controllers converge in 11.35 seconds, in average, for the cascade controller and 11.05 seconds, in average, for the pose PID controller. The vehicle is considered to converge to any point when its velocity does not exceed $0.1\ m/s$ for the remainder of the experiment.

Regarding the wide actuation swings seen in Fig.~\ref{fig:5origin_curves_angles} for the cascade controller, it is assumed that these oscillations result from the learned part of the controller trying to "subdue" the internal velocity PID controller in environments where its performance is poor. This may happen because the velocity setpoint of the internal PID controller changes rapidly, causing an increase in amplitude of these changes.

The second point to consider would be the steady-state error. This metric, combined with the settling time, is numerically evaluated in Table~\ref{tab:5curves_statistics}. Notice that, for these results, we consider the settling time of tests that did not actually settle (did not meet the maximum $0.1\ m/s$ velocity condition) to be the maximum test time, 12.5 seconds, resulting in a higher mean settling time, especially for the cascaded controller, which produces a fast oscillatory response around the target destination.

\begin{table}[tbp]
\centering
\caption{Mean and standard deviation metrics for the waypoint guidance task for 100 successful samples of each controller.}
\label{tab:5curves_statistics}
\begin{tabular}{@{}llll@{}}
\hline\noalign{\smallskip}
Controller               &
\begin{tabular}[c]{@{}l@{}} Settling time\\ (s)\end{tabular}  &
\begin{tabular}[c]{@{}l@{}} Longitudinal\\ position error\\ (m)\end{tabular} &
\begin{tabular}[c]{@{}l@{}} Vertical\\ position error\\ (m)\end{tabular} \\ \noalign{\smallskip}\hline\noalign{\smallskip}

  \textbf{Fully learned (100\%)} & $\mathbf{6.21 \; (4.38)}$ & $\mathbf{0.022 \; (0.011)}$ & $\mathbf{0.016 \; (0.008)}$ \\
  Cascade (93.5\%)  & $11.35 \; (2.23)$ & $0.057 \; (0.036)$ & $0.040 \; (0.026)$ \\
  Pose PID (95.2\%) & $11.05 \; (0.99)$ & $0.039 \; (0.025)$ & $0.027 \; (0.017)$ \\

\noalign{\smallskip}\hline
\end{tabular}
\end{table}

The larger error of the cascade controller trajectories might be caused by the oscillatory response of this controller. Despite this, the errors of the trajectories are not significantly different (sit in the same order of magnitude) and those can be improved with fine tuning of the training hyperparameters.

There is a trade-off between convergence time and success rate of the pose PID controller. By decreasing the trajectory time (thus increasing the maximum speed), the settling time can be reduced. This adjustment, on the other hand, greatly decreases the success rate of the controller due to motor saturation in a wider range of vehicle parameters and increased brashness of control actions.

%%%%

\subsection{Performance evaluation of the payload pick up task}

\begin{figure*}[tbp]
\centering
\includegraphics[width=\linewidth]{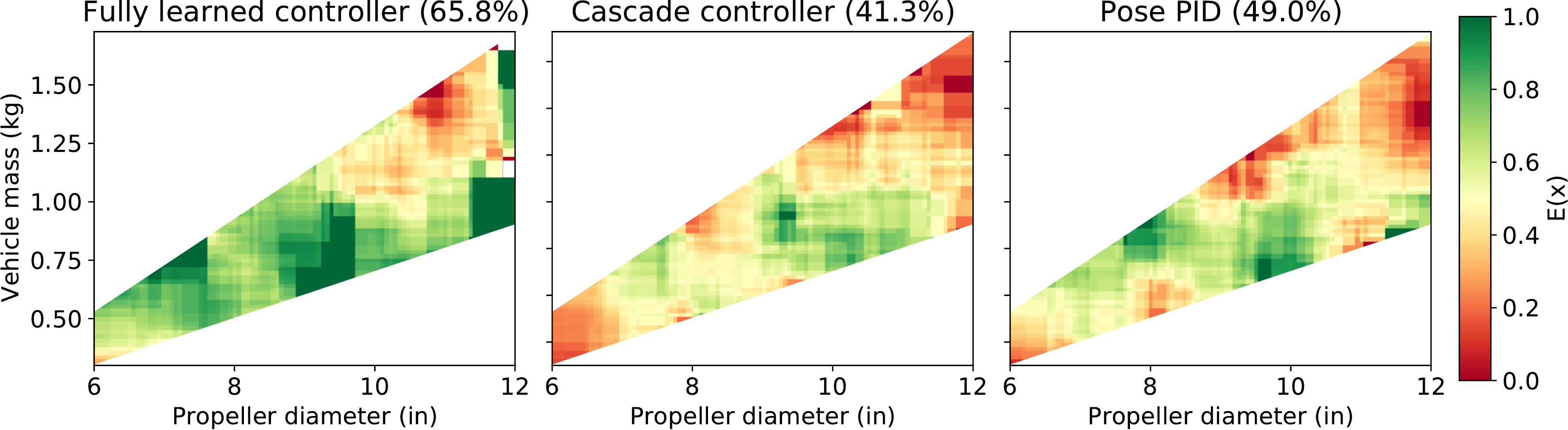}
\caption{Expected success rate for each controller in the payload pick up and drop course, by combination of quadrotor parameters.}
\label{fig:5filter_pickup}
\end{figure*}

Similarly to the waypoint guidance task, Fig.~\ref{fig:5filter_pickup} displays the expectation of a given parameter combination to perform the payload pick up and drop task successfully.

The experiments are performed using the same set of vehicle parameters as the waypoint guidance task, which is a set independent of the set used for the selection of learned controllers, described in the preamble of Sect.~\ref{sec:experiments}.

From these results, as expected from the previous results of waypoint guidance task, the pose PID controller and the cascade controller have poorer flexibility to the variation of quadrotor parameters, compared to the fully learned controller, even if the gains of their internal PID controllers are obtained using a search strategy on the same variety of vehicle parameters. All three controllers, but especially the two that rely on internal PID controllers, struggle with "boundary" vehicle parameters, that is, a set of vehicle parameters that causes it to be too heavy or to have propellers that are too small.

Another metric to evaluate the payload pick up task is the time taken to perform it, as it encompasses both the route optimality and the waypoint settling time. Note that due to the criteria of convergence to a waypoint, experiments with high position error or that take too long to reach a waypoint are filtered out (considered a failed test). Table~\ref{tab:5pick up_times} describes the mean of this metric over the experiments with a 95\% confidence range. With respect to the course completion time, the fully learned controller attains a better performance than other controllers, cascade and pose PID, these also displaying disparate completion times, the PID performing worse than other controllers.

\begin{table}[tbp]
\centering
\caption{Mean of the time taken to complete the payload pick up and drop course, for 100 successful samples of each controller, with confidence interval of 95\%.}
\label{tab:5pick up_times}
\begin{tabular}{@{}ll@{}}
\hline\noalign{\smallskip}
Controller               & \begin{tabular}[c]{@{}l@{}} Course completion\\ time (s)\end{tabular} \\ \noalign{\smallskip}\hline\noalign{\smallskip}
Fully learned (65.8\%) & $9.470 \pm 1.713$ \\
Cascade (41.3\%) & $13.364 \pm 4.370$ \\
Pose PID (49.0\%) & $25.319 \pm 1.250$ \\
\noalign{\smallskip}\hline
\end{tabular}
\end{table}

\begin{figure*}[htbp]
\centering
\includegraphics[width=\linewidth]{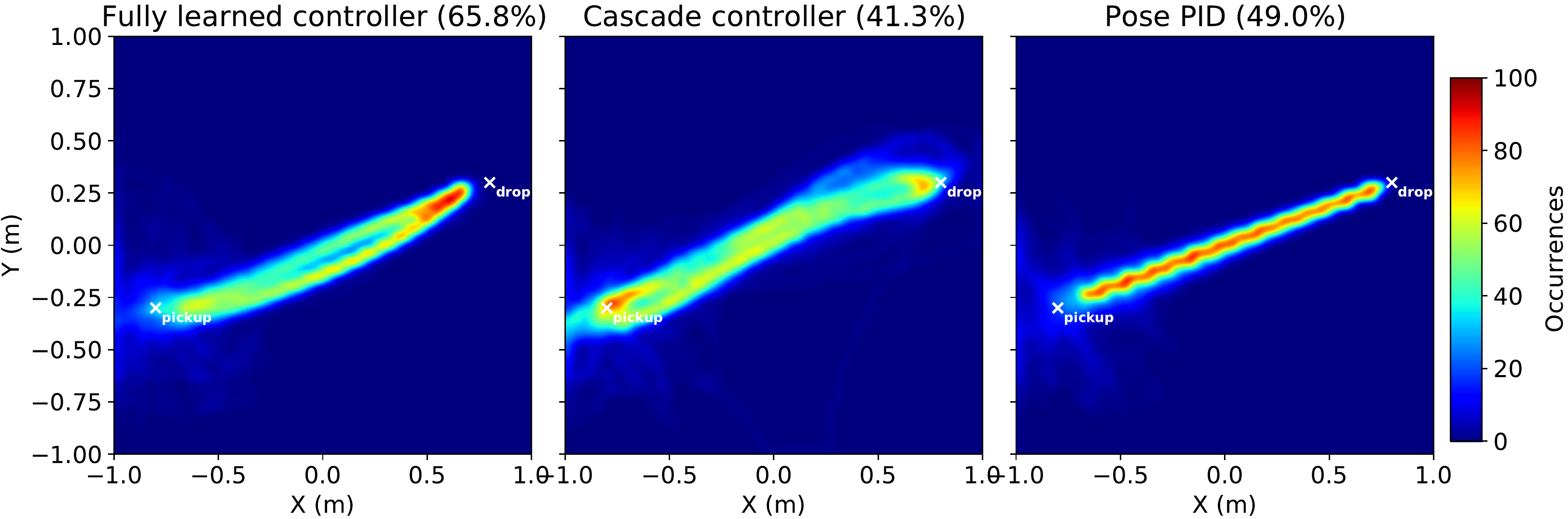}
\caption{Heatmap of the route taken by each experiment in the payload pick up and drop course, viewed from above.}
\label{fig:5heatmap_pickup_xy}
\end{figure*}

\begin{figure*}[htbp]
\centering
\includegraphics[width=\linewidth]{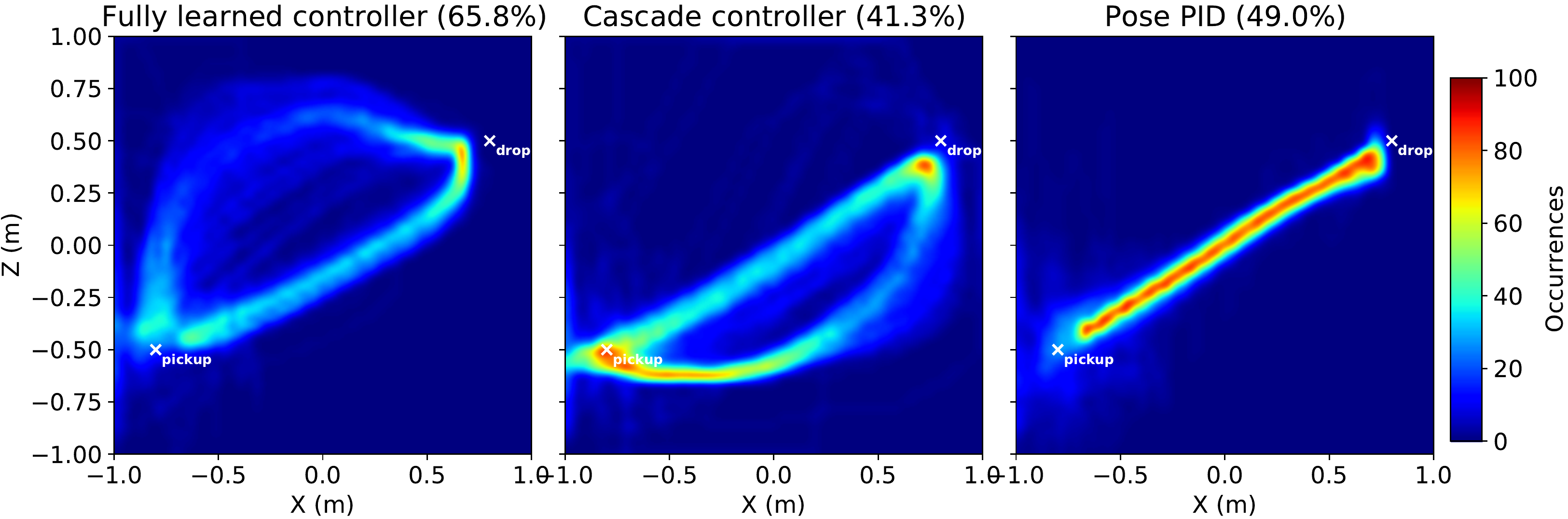}
\caption{Heatmap of the route taken by each experiment in the payload pick up and drop course, viewed from the side.}
\label{fig:5heatmap_pickup_xz}
\end{figure*}

Figs.~\ref{fig:5heatmap_pickup_xy} and \ref{fig:5heatmap_pickup_xz} show the heatmap of the compound trajectories of the quadrotor in each successful experiment. In the former map, the trajectory is viewed from above, projected in the XY plane, while in the latter map, the trajectory is viewed from the side, projected in the XZ plane.

In Fig.~\ref{fig:5heatmap_pickup_xz}, it becomes clear how the variation of mass heavily affects the trajectory generated by the evaluated controllers, as evidenced by the range of different trajectories generated, except for the pose PID controller, which, as previously observed, is able to closely (and slowly) follow the preset trajectory for any set of vehicle parameters that is able to complete the given course.

\section{Conclusion}\label{sec:conclusion}

In this work, a control policy for quadrotor navigation is developed using a Soft Actor-Critic (SAC) algorithm that maps sensory input directly into motor commands, in other words, it is learnt end-to-end. The training process for this policy incorporates domain randomization, which means that quadrotor parameters, such as mass and vehicle size, as well as initial poses, are randomly sampled for each episode. The controller is, then, evaluated in a payload pick up course, designed to attest its robustness to a number of disturbances: not only the quadrotor that is used to navigate the course is initialized with random parameters, the interaction with the payload causes changes to its dynamics mid-flight. Additionally, this task incorporates sensor noise and motor command delay, which are not present during training. Evaluation is carried out primarily considering the amount of different vehicle parameters the controller is able to handle when executing the proposed task and, secondarily, the time taken to do so.

In the proposed task, the controller is able to outperform a PID controller designed based on the \textit{Ardupilot} PID control stack, with gains tuned using a CMA evolutionary search strategy in the same set of varying quadrotor parameters. By incorporating the randomization of environment parameters in the training process, the resulting controller is also able to outperform a second controller which is trained with unchanging vehicle parameters, in the same task. Finally, using the same evaluation task, we show that an RL-based controller does not need to depend on an underlying PID controller to interact with the quadrotor: by employing an end-to-end learning strategy, the resulting controller is able to outperform a controller that is designed for high-level navigation commands only.

\subsection{Future work}

The evident next step of this work would be to investigate these controllers in a real-life quadrotor. As discussed in some of the related works, the main challenge when taking this step, assuming an adequate testing platform is readily available, would be dealing with sensor noise and deviation, especially if an inertial frame position sensor is required. It is expected that the versatility of the learned controller for different vehicle parameters provided by the domain randomization of the training environment and its robustness to unpredictable circumstances smooths out this transition, as it is very difficult to accurately model a real quadrotor in simulation.

In this work, we developed parameter-agnostic waypoint guidance controllers for plus-shaped quadrotors only. We suggest that such a controller could be modified to also account for geometrical variations of drones, such as quadrotors with movable arms, or picking up an off-centered payload, or even dealing with damaged parts mid-flight. This agnosticism can also prove useful for various multitasking applications, not only restricted to payload carrying, but any activity that imposes sudden changes in vehicle parameters.

The controller proposed in this work competes directly with the purpose of traditional robust and adaptive controllers. Even with the advantage of end-to-end learning provided by RL methods contributing to a faster development and deployment, adaptive controllers may also be an alternative for at least the waypoint guidance and payload pick up tasks proposed in this work.

%%%%%%%%%%%%%%%%%%%%%%%%%%%%%%%%%%%%

\bibliography{manuscript}

\end{document}